\documentclass[letterpaper, 10 pt, journal, twoside]{IEEEtran}
\usepackage{amsmath,amsfonts}
\usepackage{algorithmic}
\usepackage{array}
\usepackage[caption=false,font=normalsize,labelfont=sf,textfont=sf]{subfig}
\usepackage{textcomp}
\usepackage{stfloats}
\usepackage{url}
\usepackage{verbatim}
\usepackage{graphicx}
\usepackage{xurl}  

\usepackage{xspace}
\usepackage{cancel}

\usepackage[numbers]{natbib}
\usepackage{multicol}
\usepackage{todonotes}  
\usepackage{xspace}
\usepackage{placeins}
\usepackage{tcolorbox} 
\usepackage{layouts}
\usepackage{booktabs}
\usepackage{multirow}
\usepackage{makecell}
\usepackage{array}
\usepackage{gradient-text}
\usepackage{amssymb}
\usepackage{comment}
\usepackage{acronym}
\usepackage{lipsum}
\usepackage[ruled,vlined,linesnumbered]{algorithm2e}

\usepackage[dvipsnames]{xcolor} 
\usepackage[table]{xcolor}
\usepackage{colortbl}
\usepackage[bookmarks=true]{hyperref}
\usepackage{capt-of}

\usepackage{siunitx}     

\definecolor{flodarkpurple}{rgb}{0.288,0.1196,0.7}
\definecolor{amber}{rgb}{1.0, 0.75, 0.0}
\definecolor{orange}{rgb}{1.0, 0.58, 0.0}
\definecolor{green}{rgb}{0.0, 0.725, 0.25}
\definecolor{blue}{rgb}{0.047, 0.365, 0.647}
\definecolor{red_arrow}{rgb}{0.298, 0.686, 0.314}
\definecolor{green_arrow}{rgb}{1.0, 0.322, 0.322}
            
\definecolor{tablegray}{gray}{0.4} 

\definecolor{purple}{rgb}{0.29019608, 0.1372549 , 0.46666667}
\definecolor{light_blue}{rgb}{0.54901961, 0.77254902, 0.89019608}
\definecolor{pink}{rgb}{0.6627450980392157, 0.09803921568627451, 0.15294117647058825}
\definecolor{teal}{rgb}{0.07450980392156863, 0.5137254901960784, 0.2784313725490196}
\definecolor{mustard}{rgb}{0.94117647, 0.77254902, 0.44313725}

\newcommand{\mb}[1]{\mathbf{#1}}
\DeclareMathOperator*{\argmax}{arg\,max}

\newcommand{\given}{\,|\,}
\newcommand{\Mper}{\mathcal{M}_{\mathrm{P}}}
\newcommand{\Memg}{\mathcal{M}_{\mathrm{E}}}

\newcommand{\coolName}{Pre\-dict\-ive\-Graphs\xspace}
\newcommand{\coolPerpetua}{Perpetua$^*$\xspace}

\newcommand{\tinyvar}[1]{\hspace{1pt}\scriptsize{$\pm$\hspace{1pt}#1}}

\acrodef{AR}[AR]{abort rate}
\acrodef{LLM}[LLM]{large language model}
\acrodef{VLM}[VLM]{vision-language model}
\acrodef{MLLM}[MLLM]{multimodal language model}
\acrodef{MAE}[MAE]{mean absolute error}
\acrodef{BAC}[B-Acc]{balanced accuracy}
\acrodef{AIC}[AIC]{Akaike information criteria}
\acrodef{RAG}[RAG]{retrieval augmented generation}
\acrodef{CG}[CG]{ConceptGraphs}
\acrodef{CDF}[CDF]{cumulative distribution function}
\acrodef{PDF}[PDF]{probability density function}
\acrodef{PMF}[PMF]{probability mass function}
\acrodef{EM}[EM]{expectation-maximization}
\acrodef{MAP}[MAP]{Maximum a Posteriori}
\acrodef{SLAM}[SLAM]{simultaneous localization and mapping}
\acrodef{SR}[SR]{success rate}
\acrodef{SPL}[SPL]{success weighted by path length}
\acrodef{DIST}[DIST]{navigated distance}
\acrodef{PE}[PE]{persistence estimation}

\newtcolorbox[auto counter]{toolBox}[3][]{%
    colback=light_blue!5!white,
    colframe=light_blue!75!black,
    title=\textbf{Tool~\thetcbcounter: #2}, 
    label={#3}, 
    #1 
}

\newtcolorbox[auto counter]{promptBox}[3][]{%
    colback=gray!5!white,
    colframe=black!75,
    boxrule=1pt,
    arc=2mm,
    left=5pt, right=5pt, top=5pt, bottom=5pt,
    parskip=1em,
    title=\textbf{Prompt~\thetcbcounter: #2}, 
    label={#3}, 
    #1
}
\hypersetup{
  colorlinks=true,
  linkcolor=flodarkpurple,
  citecolor=teal,
  urlcolor=blue
}

\newcommand{\authorhref}[3][flodarkpurple]{\href{#2}{\color{#1}{#3}}}

\renewcommand{\baselinestretch}{.98}
\providecommand{\footref}[1]{\textsuperscript{\ref{#1}}}

\begin{document}

\begin{minipage}{\textwidth}

\begin{center}
\vspace*{\fill}
\textcopyright 2026 IEEE.  Personal use of this material is permitted.  Permission from IEEE must be obtained for all other uses, in any current or future media, including reprinting/republishing this material for advertising or promotional purposes, creating new collective works, for resale or redistribution to servers or lists, or reuse of any copyrighted component of this work in other works. 
\vspace*{\fill}
\end{center}

\hfill

This paper has been accepted for publication in the \textit{IEEE Robotics and Automation Letters}. \\ Please cite this paper as:

\begin{verbatim} 
@article{saavedra2026predictive,
	title        = {Predictive Spatio-Temporal Scene Graphs for Semi-Static Scenes},
	author       = {Saavedra-Ruiz, Miguel and Gauthier, Charlie and Gupta, Kumaraditya 
                and Shahfar, Shima and Ellis, Kirsty and Parkison, Steven and 
                Paull, Liam},
	year         = 2026,
	journal      = {IEEE Robotics and Automation Letters},
	volume       = 11,
	number       = 12,
	pages        = {14524--14531},
	doi          = {10.1109/LRA.2026.3746135},
}    \end{verbatim} 
\end{minipage}
\thispagestyle{empty}
\clearpage
\setcounter{page}{1}

\title{Predictive Spatio-Temporal Scene Graphs for Semi-Static Scenes}

\author{
\authorhref{https://mikes96.github.io}{Miguel Saavedra-Ruiz}$^{*,1,2}$,
\authorhref{https://charliegauthier.ca/}{Charlie Gauthier} $^{*,1,2}$, 
\authorhref{https://www.kumaradityag.com/}{Kumaraditya Gupta}$^{1,2}$,
\authorhref{https://scholar.google.com/citations?user=kLfWoIcAAAAJ&hl=en}{Shima Shahfar}$^{1,2}$,
\authorhref{https://www.linkedin.com/in/kirsty-ellis-943b8245}{Kirsty Ellis}$^{1,2}$, \\
\authorhref{https://scholar.google.com/citations?user=0erd9rIAAAAJ&hl=en}{Steven Parkison}$^{3}$ and
\authorhref{https://liampaull.ca/}{Liam Paull}$^{1,2,4}$
\thanks{$*$Denotes equal contribution.}
\thanks{$^{1}$Department of Computer Science and Operations Research, Université de Montréal, Montréal, QC, Canada.}
\thanks{$^{2}$Mila - Quebec AI Institute, Montréal, QC, Canada.}
\thanks{$^{3}$Hydro-Québec Research Institute, Varennes, QC, Canada.}
\thanks{$^{4}$Canada CIFAR Chair.}
\thanks{\textit{Corresponding author: Miguel Saavedra-Ruiz} -- \textit{email}: {\tt\footnotesize miguel.angel.saavedra.ruiz@umontreal.ca}}
\thanks{Accepted for publication in IEEE Robotics and Automation Letters (RA-L).}
}

\makeatletter
\let\@oldmaketitle\@maketitle
\renewcommand{\@maketitle}{\@oldmaketitle
\centering
\vspace{1em}
\includegraphics[width=\linewidth,trim={0.5cm 0 0.01cm 0},clip]{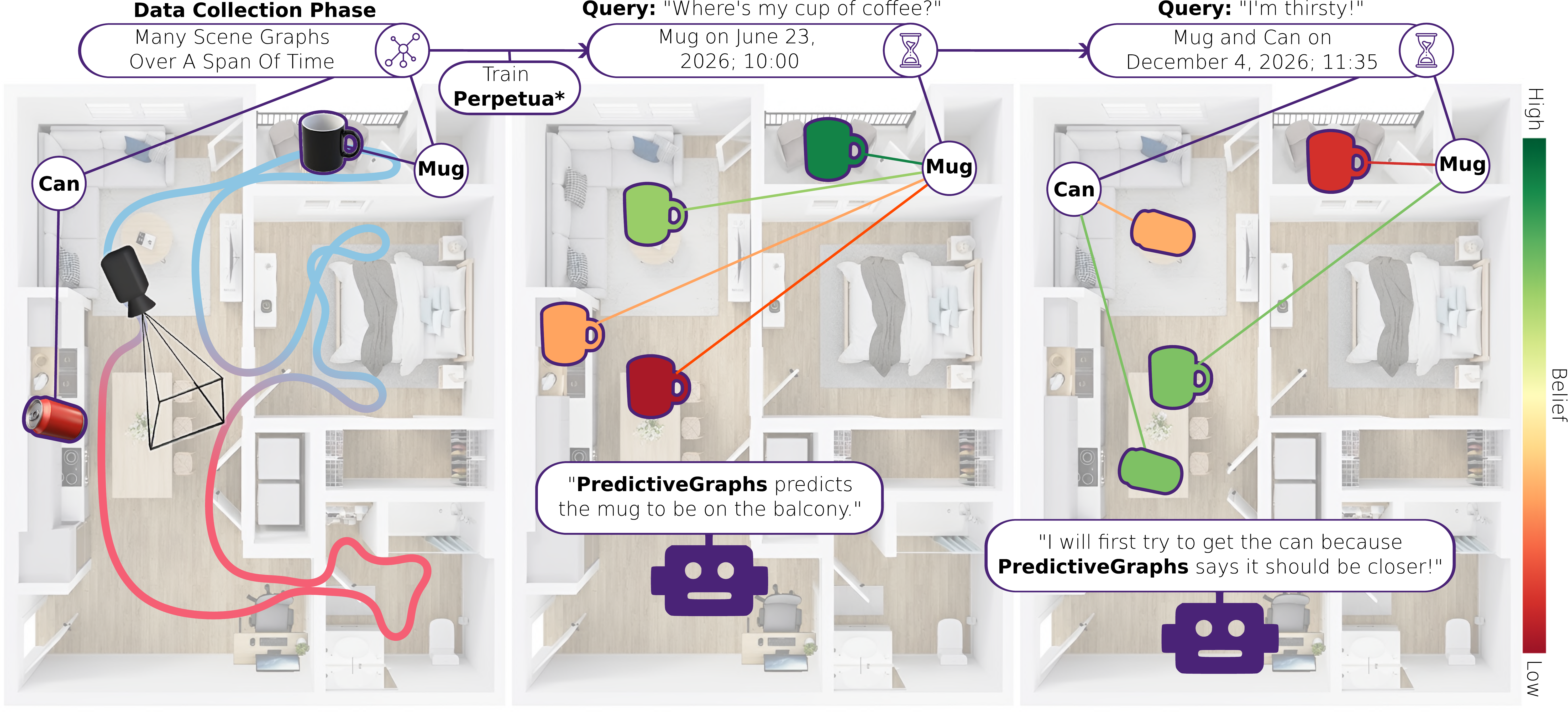}
\vspace{-2.0em}
\setcounter{figure}{0} 
\captionof{figure}{
\footnotesize
\textbf{\coolPerpetua} augments open-vocabulary \textcolor{purple}{scene graph representations} built by state-of-the-art perception algorithms to produce \textbf{\coolName}. The resulting representation supports predictive tempo-spatio-semantic queries. 
\vspace{-1em}
}
\label{fig:banner}
}
\makeatother

\maketitle

\begin{abstract}
We have seen tremendous recent progress in our ability to build ``spatio-semantic'' representations that enable robots to perform complex reasoning across geometry and semantics. However, the vast majority of these methods lack any ability to perform reasoning across time. This is a desirable property in situations where a robot repeatedly observes an environment where instances may change in between observations, but in a structured way. Consider as an example a home environment where the location of a mug typically moves from the cupboard to a countertop to the sink and then back to the cupboard on a daily basis. We should be able to learn this cyclic behavior and use it to predict the state of the mug in the future. In this work, we propose a method that is able to perform this type of tempo-spatio-semantic reasoning. Underpinning the method is a filter, \coolPerpetua, that performs Bayesian reasoning on the states of the environment that are observed over time. This filter is integrated within a 3D scene graph structure that we call \coolName, where nodes represent objects and edges function as \coolPerpetua filters encoding spatio-semantic relationships. We validate the method in both simulation and real-world dynamic navigation tasks, where our real-world experiments consist of an environment that is undergoing semi-static changes at a bi-hourly frequency over a period of three weeks. In both settings, we demonstrate that our method outperforms baselines in predicting future environment states, even in the presence of distributional shifts.\looseness-1
\end{abstract}

\section{Introduction}

Achieving long-term operation in dynamic environments remains a fundamental challenge in robotics, spanning nearly three decades of active research~\cite{cyrill2005nonStatic, Biber2005DynamicMF}. Solving this problem is key to unlocking truly autonomous agents capable of reliable operation in an ever-changing world without constant human intervention~\cite{slam-handbook}. A fundamental challenge lies in  the complex nature of \emph{semi-static changes}, where objects often appear or disappear between observations, a challenge exacerbated by partial, noisy sensor data in real-world deployments~\cite{schmid2024khronos}. 

Effective handling of such dynamics requires robots  to reason across temporal dimensions: 
leveraging the past to contextualize observations, tracking the present state of the world, and predicting the future to anticipate changes and enable informed decision-making. However, while many state-of-the-art methods rely on change-detection to maintain an up-to-date map~\cite{ schmid2022panoptic}, they typically maintain only a hypothesis about the \emph{present}. By discarding historical data to satisfy memory constraints, 
these approaches sacrifice the temporal information necessary for predicting environmental evolution.\looseness-1

To address this gap, recent approaches have begun exploiting \emph{past} temporal information to better inform map updates~\cite{schmid2024khronos}, or ground complex queries about the past in spatial maps~\cite{anwar2025remembr, yuan2026star} (e.g., ``where were my keys four hours ago?''). While these methods successfully reason about \emph{past} and \emph{present}, reasoning about the \emph{future} is largely unexplored.\looseness-1

Yet, most real-world scenes exhibit semantically consistent patterns across time that we should be able to learn, e.g., a door consistently open from 9 to 5, or an office building being empty on weekends. Persistence estimators~\cite{Rosen2016Towards, nobre2018online} are methods designed to learn and predict such semi-static dynamics. However, existing approaches face a fundamental trade-off: filtering-based methods like Perpetua~\cite{saavedra2025perpetua} can track changes in real-time by updating beliefs as new observations arrive but have limited long-term forecasting capabilities, while prediction-based estimators~\cite{krajnik2017fremen, wang2024arma} excel at long-horizon forecasting but lack adaptation capabilities.

In this paper, we bridge these gaps by first proposing \coolPerpetua, a persistence estimator that extends Perpetua~\cite{saavedra2025perpetua} with tempered Bayesian model averaging~\cite{raftery2010online}. Specifically, we use a switching prior from either environment-specific observations~\cite{krajnik2017fremen} or a \ac{LLM}~\cite{comanici2025gemini} to improve Perpetua's persistence estimates over long horizons. 
As a result, \coolPerpetua maintains the core strengths of its predecessor while overcoming its predictive limitations.

We then couple \coolPerpetua\ with a scene graph (ConceptGraphs~\cite{gu2024conceptgraphs}) to create \coolName, a map with extended predictive capabilities for object-level semi-static changes over long horizons. Specifically, we attach persistence estimators to objects in the map to track their temporal patterns across static annotated locations. Our key insight is that in human-centric and structured settings (e.g., warehouses), objects exhibit temporal structure driven by routine rather than randomness~\cite{schmid2022panoptic, yugay2025gaussian, qian2023povslam}. To render \ac{PE} tractable, we assume that semi-static objects have a finite number of discrete states that we call ``receptacles''. Receptacles are annotated a priori, while object detection and object-receptacle association remain open-vocabulary. Combining persistence estimators with scene graphs allows these map representations to (1) break the staticity assumption and accurately model semi-static objects, (2) predict \emph{future} object states, including relocation or removal from the environment, and (3) remain continuously updated based on learned temporal patterns.
We demonstrate the utility of this approach through a dynamic object navigation task.\looseness-1

In summary, our contributions are: 

\begin{enumerate}
    \item \textbf{\coolPerpetua}, a persistence estimator that extends Perpetua with a switching prior via tempered Bayesian model averaging, enabling long-horizon predictions while preserving real-time adaptability; 
    \item \textbf{\coolName}, a tempo-spatio-semantic scene graph that integrates \coolPerpetua to model object-level semi-static dynamics; and 
    \item A \textbf{long-term dataset} of semi-static dynamics spanning three weeks in a laboratory setting, with bi-hourly observations during operational hours\footnote{\label{fn:supp}Webpage: \url{https://montrealrobotics.ca/predictive-graphs}.}.
\end{enumerate}

We validate our approach on both simulation data and a novel real-world dataset capturing semi-static dynamics in a laboratory over three weeks. Through these experiments, we demonstrate that our method can address a new type of spatio-temporal queries that previous methods could not address by propagating our map representation forward in time.

\section{Related Work}

\textbf{Persistence Estimation.} \acs{PE} models and predicts the temporal dynamics of semi-static features~\cite{Rosen2016Towards}. This is distinct from spatio-temporal \ac{SLAM} methods~\cite{schmid2024khronos, behrens2025lost}, which track but do not forecast 
changes. Existing approaches include \textit{predictive methods}, such as FreMEn~\cite{krajnik2017fremen} and ARMA~\cite{wang2024arma}, which excel at long-horizon forecasting via frequency-domain or autoregressive modeling but struggle with distributional shifts, and \textit{filtering methods}, such as the persistence filter~\cite{Rosen2016Towards} and Perpetua~\cite{saavedra2025perpetua}, which adapt to real-time changes but lack long-term predictive power. \coolPerpetua bridges this gap by extending Perpetua with a switching prior via tempered Bayesian model averaging.

While several works have integrated \acs{PE} with spatial representations~\cite{guizilini2019hilbert, waqas2026lilo}, these methods operate on low-level geometric features, lacking the semantic understanding necessary for symbolic reasoning. Similarly, graph neural network-based methods for temporal link prediction~\cite{kurenkov2023modeling, homer} share our goal of forecasting scene graph changes. However, unlike \coolPerpetua, these approaches do not recursively adapt to new observations.\looseness-1 

\textbf{Scene Graphs.} Open-vocabulary 3D scene graphs, such as ConceptGraphs~\cite{gu2024conceptgraphs}, HOV-SG~\cite{werby2024hierarchical}, and Clio~\cite{maggio2024clio}, have evolved from closed-vocabulary models  by embedding \acp{VLM} representations. Recent methods exploit temporal information to inform map updates~\cite{yugay2025gaussian, liu2025dynamem} or integrate it into map optimization~\cite{schmid2024khronos, qian2023povslam}. Spatio-temporal representations have incorporated \acs{VLM}s with \ac{RAG}~\cite{lewis2020retrieval} to ground queries about past events in spatial maps~\cite{anwar2025remembr, yuan2026star}. Although scene graphs such as DualMap~\cite{jiang2025dualmap}, DynaMem~\cite{liu2025dynamem}, and OpenIn~\cite{tang2025openin} model semi-static environments, and others predict graph changes~\cite{homer, kurenkov2023modeling}, \coolName is the first to add predictive capabilities to a scene graph with open-vocabulary object recognition.\looseness-1 

\section{Problem Statement}
\label{sec:problem}

Following the problem formulation of Saavedra-Ruiz et al.~\cite{saavedra2025perpetua}, we consider an agent operating in an environment undergoing semi-static changes, where objects may appear or disappear between observations. Similar to previous semantic scene graph work~\cite{jiang2025dualmap, gu2024conceptgraphs, werby2024hierarchical}, we assume access to posed RGB-D frames $\mathcal{I} = \{I_{t_1}, I_{t_2}, \dots, I_{t_N}\}$ collected over multiple sessions at times \(\{t_i\}_{i=1}^N \in [0, \infty)\). These observations are used to construct and maintain a metric-semantic scene graph $\mathcal{G}_{t_N} = \langle \mathcal{V}_{t_N}, \mathcal{E}_{t_N} \rangle$ up to time $t_N$, where $\mathcal{E}_{t_N}$ denotes the edges, and $\mathcal{V}_{t_N} = \mathcal{O}_{t_N}^{S} \cup \mathcal{O}_{t_N}^{R} \cup \mathcal{O}^{BG}$ comprises semi-static objects, receptacles, and background elements, respectively.

Each semi-static object $o_j \in \mathcal{O}_{t_N}^{S}$ is connected via edges to the set of receptacles $\mathcal{R}_{j} \subseteq \mathcal{O}^R_{t_N}$ where it has been previously observed. Every edge $(o_j, o_k) \in \mathcal{E}_{t_N}$ carries a binary persistence variable $X_t^{j,k} \in \{0,1\}$, that takes the value $1$ if object $o_j$ is present in receptacle $o_k$ at time $t \in [t_N, \infty)$, and $0$ otherwise. Therefore, given a (potentially) outdated scene graph $\mathcal{G}_{t_N}$, the goal is to answer text queries at time $t > t_N$ (e.g., ``where is my coffee mug?'') by inferring the posterior over all edges $X_t^{j,k}$ and identifying the most likely receptacle containing the target object, or flagging the object as absent.\looseness-1

\section{\coolPerpetua}

In this section we present our persistence estimator that we call \coolPerpetua, which models the presence or absence of a specific feature in the environment over time. During the exposition here, we will consider  a single environment feature.

\subsection{Preliminaries: Perpetua}
\label{sec:perpetua}

Perpetua~\cite{saavedra2025perpetua} is a persistence estimator that extends the persistence filter~\cite{Rosen2016Towards} with a mixture formulation to capture multiple persistence hypotheses. Since Perpetua models individual features at single locations, we omit receptacle indices and let $X_t$ denote the binary feature state. Let \(\{Y_{t_i}\}_{i=1}^N\) with $Y_{t_i} \in \{0, 1\}$ denote random variables modelling noisy presence observations obtained from an arbitrary feature detector.

Specifically, the mixture of persistence filters works by estimating multiple persistence hypotheses over survival time, $T$, which characterizes how long a feature exists before disappearing. The generative model is defined as:\looseness-1
\begin{equation}\label{eq:perpetua_model}
    \begin{aligned}
    T \mid C &= l \sim p_{l}(\cdot \mid C = l), \\
    X_t \mid T &= 
    \begin{cases} 
    1, & t \leq T, \\
    0, & t > T
    \end{cases} \\
    Y_t \mid X_t &\sim p_{Y_t}(\cdot \mid X_t);
    \end{aligned}
\end{equation}
\noindent
where $C \in \{1,\dots, L\}$ is a categorical variable denoting the mixture components, with prior probability $\pi_l = p(C = l)$ satisfying $\sum_{l=1}^L \pi_l = 1$. The function \(p_l\) denotes the conditional density function of the survival time $T \in [0, \infty)$ given the $l$-th mixture component. The binary variable $X_t \in \{0, 1\}$ models feature presence at time $t$. Note that the events $X_t = 1$ and $T \geq t$ are equivalent. Finally, the measurement model $p_{Y_t}$ accounts for observation noise and is characterized by the probability of false negative \( P_M = p(Y_t = 0 | X_t = 1) \), and false positive \(P_F = p(Y_t = 1 | X_t = 0)\). From now on, we adopt the notation: $x_t \triangleq (X_t = 1)$ and $c_l \triangleq (C = l)$.

Given a sequence of noisy observations $\mathcal{Y}_{1:N} \triangleq \{y_{t_i}\}_{i=1}^N$, the mixture of persistence filter's posterior at time $t \geq t_N$ is:
\begin{equation}
    \label{eq:perpetua_posterior}
    \sum_{l=1}^L w_l \; p(x_t \given c_l,  \mathcal{Y}_{1:N}) = \sum_{l=1}^L w_l \; \frac{p(\mathcal{Y}_{1:N} \given x_t)p_l(x_t \given c_l)}{p(\mathcal{Y}_{1:N} \given c_l)},
\end{equation}
\noindent
with $p(x_t \given c_l,  \mathcal{Y}_{1:N})$ denoting the conditional posterior of each mixture, $w_l \triangleq p(c_l \given \mathcal{Y}_{1:N})$ the posterior weights, and $p(\mathcal{Y}_{1:N} \given x_t)$, $p_l(x_t \given c_l)$ the measurement likelihood and prior, both updated at each new observation as 
\begin{align}
    \label{eq:perpetua_likelihood}
    p(\mathcal{Y}_{1:N} \mid x_t) &= \prod_{i=1}^NP_M^{1 - y_{t_{i}}} (1 - P_M)^{y_{t_{i}}}, \\
    \label{eq:perpetua_prior}
    p_l(x_t \mid c_l) &= 1 - F_l(t) = 1 - \int_0^t p_l(\tau|c_l)d\tau.
\end{align}

The conditional evidence $p(\mathcal{Y}_{1:N} \given c_l)$ decomposes as a sum over disjoint survival intervals and thus admits a recursive update~\cite{saavedra2025perpetua}. Writing
$\ell_N \triangleq p(\mathcal{Y}_{1:N} \given x_{t})$ for
\eqref{eq:perpetua_likelihood} and $\Delta F_N \triangleq F_l(t_N) - F_l(t_{N-1})$,
\begin{equation}
    \label{eq:perpetua_evidence}
    H_N = P_F^{y_{t_N}} (1 - P_F)^{1-y_{t_N}}
          \bigl( H_{N-1} + \ell_{N-1}\, \Delta F_N \bigr),
\end{equation}
\noindent
so that $p(\mathcal{Y}_{1:N} \given c_l) = H_N + \ell_N [1 - F_l(t_N)]$, with
$H_N$ acting as a recursive accumulator, facilitating the efficient update of the conditional evidence. Once the conditional posterior,  $p(x_t \given c_l,  \mathcal{Y}_{1:N})$, is computed for all mixtures, the posterior weights can be recovered via 
\begin{equation}
    \label{eq:perpetua_weights}
    w_l = \frac{p(\mathcal{Y}_{1:N}, c_l)}{p(\mathcal{Y}_{1:N})} = \frac{p(\mathcal{Y}_{1:N} \mid c_l)p(c_l)}{\sum_{l=1}^L p(\mathcal{Y}_{1:N} \mid c_l)p(c_l)}.
\end{equation}

Finally, instead of using the marginalized mixture posterior from \eqref{eq:perpetua_posterior}, the final prediction is based on the dominant component (i.e., the component with the maximum posterior weight):\looseness-1
\begin{align}
    \label{eq:perpetua_mode}
    p(x_t \mid c_{l^*}, \mathcal{Y}_{1:N}) \;\; \text{s.t.} \;\; l^* = \argmax_{l} p(c_l \mid \mathcal{Y}_{1:N}).
\end{align}
To capture feature reappearance, Perpetua introduces a complementary mixture of \emph{emergence} filters, derived analogously to the persistence model in~\eqref{eq:perpetua_model}. To handle both appearance and disappearance, the persistence and emergence models are coupled via a state machine, with switching governed by a heuristic threshold on~\eqref{eq:perpetua_mode}. Perpetua has two notable limitations: the heuristic switching requires costly simulation steps to detect state changes, and long-horizon predictions are limited to periodic patterns in the absence of observations (see Fig.~\ref{fig:pred_odd}), which may diverge from true feature dynamics.

\subsection{Method}

\label{sec:perpetua_star}

The main contribution of \coolPerpetua lies in replacing the heuristic state machine from  Perpetua~\cite{saavedra2025perpetua} with tempered Bayesian model averaging~\cite{murphy2022pml1}, which provides a probabilistic basis for weighting the emergence and persistence models. In particular, our proposed model averaging is informed by a new switching prior, allowing the estimator to leverage environment-specific knowledge or \acp{LLM}.

We propose a switching prior, $f(t): [0, \infty) \rightarrow [0, 1]$, which can take several forms, such as:
\begin{equation}
    \label{eq:priors}
    f_{\text{FreMEn}}(t) \hspace{-0.5mm} = \sigma \left( \mathcal{F}^{-1}[\mathcal{P}(\omega)](t) \right) \; \text{or} \;\; f_{\text{LLM}}(t) = \hspace{-0.5mm} \sum_{r=1}^{R} a_r  \mathbb{I}(t \in \mathcal{T}_r).
\end{equation}
\noindent
Here, $f_{\text{FreMEn}}$ applies the inverse Fourier transform $\mathcal{F}^{-1}$ to coefficients $\mathcal{P}(\omega)$ with $\sigma(x) = \min ( \max (x, 0), 1)$ \cite{krajnik2017fremen}; and $f_{\text{LLM}}$ defines a piecewise constant function using the indicator function $\mathbb{I}(\cdot)$. In the former, data is used to learn an informed prior for the target environment. In the latter, an \acs{LLM} can be used in the absence of data to obtain a commonsense prior.\looseness-1

\begin{figure}[t]
    \centering
    \includegraphics[trim={0 0 0 1.98mm}, clip, width=\columnwidth]{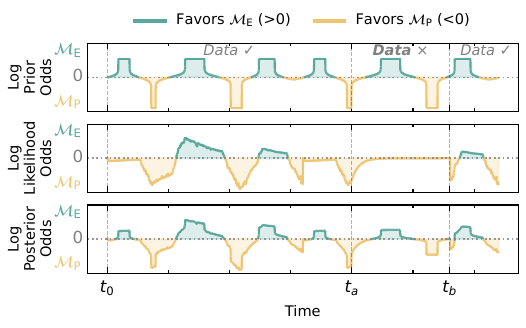}
    \vspace{-25px}
    \caption{\textbf{\coolPerpetua Tempered Bayesian Model Averaging}: 
    Positive values favor the emergence model, while negative values favor the persistence model. The data gap ($t_a$ to $t_b$) causes the model likelihood to decay, leaving the posterior influenced only by the prior until observations resume.
    \vspace{-2em}
    }
\label{fig:model_selection}
\end{figure}

Given this prior, we define $\mathcal{M}$ as a binary random variable selecting between the persistence model $\Mper$ and emergence model $\Memg$ (both defined as in \eqref{eq:perpetua_model}), with switching probability $p_t(\mathcal{M} = \Memg) = f(t)$ and $p_t(\mathcal{M} = \Mper) = 1 - f(t)$, governing which model is more likely active at time $t$. Unlike Perpetua~\cite{saavedra2025perpetua}, we update both models using all available observations $\mathcal{Y}_{1:N}$. We then employ tempered Bayesian model averaging~\cite{murphy2022pml1} to weight the two models, treating the marginal evidence (denominator in \eqref{eq:perpetua_weights}) as the model likelihood $p(\mathcal{Y}_{1:N} \given \mathcal{M})$ and combining it with the prior $p_t(\mathcal{M})$:
\begin{equation}
    \label{eq:perpetua_model_selection}
    p(\mathcal{M} \mid \mathcal{Y}_{1:N}) \propto p(\mathcal{Y}_{1:N} \mid \mathcal{M})^{\alpha(t)}p_t(\mathcal{M}).
\end{equation}

Since the model likelihood is only updated upon receiving new observations, we introduce an exponential annealing term $\alpha(t) = \exp(-\alpha_0 (t - t_N))$, where $\alpha_0 > 0$ is a decay rate and $t_N$ is the last observation time. This term gradually tempers the likelihood of the model such that in the prolonged absence of data, model averaging becomes increasingly governed by the prior $p_t(\mathcal{M})$. With the switching posterior in \eqref{eq:perpetua_model_selection} computed, \coolPerpetua generates predictions for any $t \geq t_N$ using a weighted average of the persistence and emergence models:
\begin{equation}
    \label{eq:perpetua_final_pred}
    p(x_t \given \mathcal{Y}_{1:N}) = \hspace{-4mm} \sum_{m \in \{\Memg, \Mper\}} \hspace{-4mm} p(m \given \mathcal{Y}_{1:N})  p(x_t \given c_{l^*}, \mathcal{Y}_{1:N}, m),
\end{equation}
\noindent
with predictions $p(x_t \given c_{l^*}, \mathcal{Y}_{1:N}, m)$ computed via \eqref{eq:perpetua_mode} using the model parameters $m$, and dominant mixture component $c_{l^*}$. For rapid adaptation, we apply a forgetting factor $\gamma \in [0,1]$ to the accumulated measurement likelihood in~\eqref{eq:perpetua_likelihood} at every update, $p(\mathcal{Y}_{1:N} \given x_t) = p(\mathcal{Y}_{1:N-1} \given x_{t})^{\gamma} \, P_M^{1-y_{t_N}} (1-P_M)^{y_{t_N}}$ and the recursive accumulator $H_N$ in~\eqref{eq:perpetua_evidence}. As in non-stationary model averaging~\cite{raftery2010online}, this recursion tempers each measurement by a geometrically decaying exponent, entering the evidence update in~\eqref{eq:perpetua_evidence} through every likelihood term. Setting $\gamma = 1$ recovers \eqref{eq:perpetua_likelihood} exactly, while $\gamma < 1$ bounds the accumulated evidence so that recent observations
dominate. Figure~\ref{fig:model_selection} shows how the switching prior (top) and model likelihood (middle) in \eqref{eq:perpetua_model_selection} interact to yield the final model hypothesis (bottom). A step-by-step example of how all \coolPerpetua components interact is given in Fig.~\ref{fig:fig_example} (App.~\ref{app:complexityanalysis}).\looseness-1

\subsubsection{Multi-Receptacle Belief}

\coolPerpetua as described handles a single \textit{feature}: an object-receptacle pair (is the mug on the balcony?) or an environmental state (is it raining?). To handle multiple receptacles, we aggregate individual \coolPerpetua estimators to track object $o_j \in \mathcal{O}_{t_N}^{S}$ across $K$ receptacles, by assuming the factorization: $p(x_t^{j,1}, \dots, x_t^{j,K} \given \mathcal{Y}_{1:N}) = \prod_{k=1}^K p(x_t^{j,k} \given \mathcal{Y}^k)$. Here, $\mathcal{Y}^k$ denotes the observations for the $k$-th receptacle, and $p(x_t^{j,k} \given \mathcal{Y}^k)$ represents the state belief over that receptacle, computed via \eqref{eq:perpetua_final_pred}. This approximation allows an object to be assigned to multiple locations simultaneously. Empirically, conflicting high-probability edges are rare (App.~\ref{app:independence}).\looseness-1

This is sometimes desirable: perception systems suffer from a well-known ``indistinguishable objects'' failure mode~\cite{grinvald2019volumetric}, wherein different physical objects might be considered the same semantic class (e.g., three white mugs all labelled ``mug''). But this is not desirable in cases where perception is trustworthy, or where a concrete decision must be taken (e.g., a navigation system must decide which receptacle to visit). In this case we resolve this ambiguity by identifying the single most likely receptacle:\looseness-1
\begin{equation}
    \label{eq:perpetua_map}
    k^* = \argmax_k \; p(x_t^{j,k} \given \mathcal{Y}^k).
\end{equation}
Importantly, by comparing the final estimate $p(x_t^{j,k^*} \given \mathcal{Y}^{k^*})$ against a probability threshold $\delta$, this formulation determines not only the location of the object but also its potential absence from the scene (see Fig.~\ref{fig:perpetua_star} and Alg.~\ref{alg:mixture_filter} in App.~\ref{app:algorithm}). In this sense, \coolPerpetua is not merely a model averaging criterion, but a principled framework that bridges the filtering and predictive persistence estimation literature into a unified, theoretically grounded method, allowing practitioners to flexibly swap filtering and predictive (switching prior) components as needed.

\section{\coolName}

In this section, we demonstrate the utility of persistence estimators by embedding \coolPerpetua into an open-vocabulary scene graph, unlocking predictive capabilities over semi-static changes directly within the map. We call the resulting representation \coolName, a map that jointly reasons over space, semantics, and time. Similar to~\cite{gu2024conceptgraphs,werby2024hierarchical}, we construct this graph by lifting 2D region proposals from posed RGB-D frames (obtained from a \ac{SLAM} algorithm) into 3D and merging observations over time. Following the notation in Sec.~\ref{sec:problem}, a node $o_u \in \{ \mathcal{O}^S_{t_N} \cup \mathcal{O}^R_{t_N} \cup \mathcal{O}^{BG} \}$ comprises a feature embedding $e_u \in \mathbb{R}^d$ (e.g., from CLIP~\cite{radford2021learning}) to enable arbitrary natural language queries and a point cloud $\mathcal{P}_u \subset \mathbb{R}^{3}$. Each edge $(o_j, o_k) \in \mathcal{E}_{t_N}$ is augmented with a \coolPerpetua estimator tracking the persistence state of semi-static objects.

\subsection{Method}
\label{sec:pg_method}

\subsubsection{Object Classification}
\label{sec:obj_class}
Given a scene graph $\mathcal{G}_{t_N}$ constructed from observations $\mathcal{I}$ from the latest mapping session in the time interval $(t_i, t_N]$, we first classify each object node into one of three categories: semi-static $\mathcal{O}^S_{t_N}$, receptacle $\mathcal{O}^R_{t_N}$, or background $\mathcal{O}^{BG}$. 

\begin{figure}[t]
    \centering
    \includegraphics[width=0.95\columnwidth]{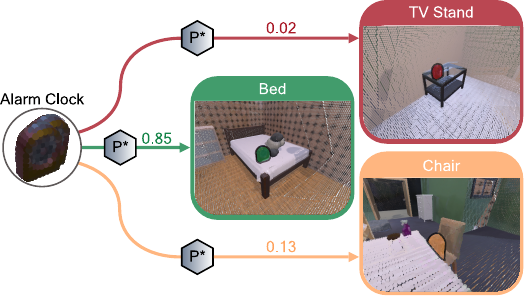}
    \vspace{-10px}
    \caption{\textbf{Edge Update:} Each semi-static object node has an associated set of \coolPerpetua estimators (one for each receptacle node it has been observed at). At prediction time, \coolPerpetua updates the belief of each edge, with {\color{teal}green} indicating the object's most likely receptacle and {\color{pink}red} its least likely one.\looseness-1
    \vspace{-2em}
    }
    \label{fig:perpetua_star}
\end{figure}

We first use CLIP to retrieve the objects most likely to be semi-static, using a text description of what a semi-static object is as the target embedding, and rank the top-$k$ candidates by similarity. An object $o_j$ is added to $\mathcal{O}^S_{t_N}$ only after \acs{LLM} verification to remove false positives. Receptacles are large, stationary objects (e.g., tables, shelves) that serve as containers for semi-static objects\footnote{Note, however, that the ``receptacle'' term can be a misnomer. \coolName might consider a door to be the semi-static ``object'', and a placeholder such as ``doorframe’’ to be its receptacle.}. Since receptacles are fixed throughout operation, we identify them using manually annotated bounding boxes (open-vocabulary receptacle detection under-segments our scenes; see App.~\ref{app:receptacle_detection}) and formulate node-to-receptacle assignment as a bipartite matching problem using 3D IoU, solved via the Hungarian algorithm. Matched receptacle nodes are added to $\mathcal{O}^R_{t_N}$; all remaining nodes are classified as background $\mathcal{O}^{BG}$.

\subsubsection{Spatial Association \& Edge History}
\label{sec:edge_history}

After classification, for each mapping session $\lambda \in \Lambda$ (with $\Lambda$ denoting the set of all mapping sessions), we determine semi-static object-receptacle associations by computing the Euclidean distance between point cloud centroids: the per-session association is $A^{(\lambda)}(o_j, o_k) = 1$ if semi-static object $o_j \in \mathcal{O}^S_{t_N}$ is assigned to receptacle $o_k \in \mathcal{O}^R_{t_N}$ in session $\lambda$, and $0$ otherwise. We then aggregate across all sessions: $A(o_j, o_k) = 1$ if $o_j$ has been observed at $o_k$ in \emph{any} session, and $0$ otherwise. This historical record forms the basis for the temporal edges of \coolName, as described next.\looseness-1

\subsubsection{Temporal Edge Formulation}
Having classified the objects and built the edge history, we construct the edge set $\mathcal{E}_{t_N}$ to encode spatio-temporal object-receptacle relationships. Unlike standard scene graphs that define sparse edges based on spatial proximity, we adopt a \emph{dense} edge formulation that connects each semi-static object to all receptacles at which it has been historically observed via a persistence variable $x_t^{j,k}$. \looseness -1

Formally, for each semi-static object $o_j \in \mathcal{O}^S_{t_N}$, we establish connectivity to the set of receptacles at which it has been observed: $\mathcal{R}_j = \{ o_k \in \mathcal{O}^R_{t_N} \mid A(o_j, o_k) = 1 \}$. Each edge between $(o_j, o_k)$ with $o_k \in \mathcal{R}_j$ has a weight $x_t^{j,k}$ given by the \coolPerpetua belief computed via~\eqref{eq:perpetua_final_pred}. This formulation yields a time-varying scene graph $\mathcal{G}_{t_N} = (\mathcal{V}_{t_N}, \mathcal{E}_{t_N})$, whose edge weights are learned from the edge history and evolve according to learned persistence dynamics, enabling semi-static object localization, as illustrated in Fig.~\ref{fig:perpetua_star}.

\subsubsection{Temporal Queries and Map Propagation}

The integration of \coolPerpetua with the scene graph enables querying the map at arbitrary future times $t > t_N$, even in the absence of new observations, by recomputing the edge weights for all semi-static object-receptacle pairs using the corresponding \coolPerpetua estimators.
For downstream tasks such as object navigation, we use the top-$k$ receptacles ranked by edge weight, providing a prioritized list of locations to visit. 

\subsubsection{\coolPerpetua Belief Updates}

As the robot navigates, \coolPerpetua estimators are updated online with new observations. For each RGB-D image $I_t$ acquired at time $t > t_N$, we construct a local scene graph $\mathcal{G}_t^{\text{local}}$ and apply the same classification procedure described in Sec.~\ref{sec:obj_class} to identify local semi-static objects $\mathcal{O}_t^{S,\text{local}}$ and receptacles $\mathcal{O}_t^{R,\text{local}}$.

We then compute a local spatial association (Sec.~\ref{sec:edge_history}) and for each local receptacle $o_k \in \mathcal{O}_t^{R,\text{local}}$ matching a known receptacle in $\mathcal{O}^R_{t_N}$, we generate measurements for the corresponding \coolPerpetua\ estimator: a presence measurement ($y_t = 1$) for each semi-static object $o_j \in \mathcal{O}_t^{S,\text{local}}$ found on $o_k$, and an absence measurement ($y_t = 0$) for each $o_j \in \mathcal{O}^S_{t_N}$ that is \emph{not} detected on $o_k$ in the local observation even though it was previously observed there. This online update mechanism enables \coolName to adapt to distribution shifts where test-time dynamics diverge from the training data.

\subsection{Embodied LLM Planning}
\label{sec:llm_planning}

For language-driven embodied navigation, we implement an LLM-based planning agent over \coolName. The agent is equipped with three tools: \emph{semantic search} via CLIP~\cite{radford2021learning} for open-vocabulary object queries, \emph{location prediction} via \coolPerpetua to estimate the environment's evolution at time $t$, and \emph{active navigation} that integrates path planning with continuous map updates (full specifications in App.~\ref{app:tools}). Together, these tools allow the agent to identify query objects, form a hypothesis of where they are, and navigate toward candidate receptacles until they are found.

\section{Evaluation}

\begin{figure}[t]
    \centering
    \includegraphics[trim={0 0 0 1.98mm}, clip, width=\columnwidth]{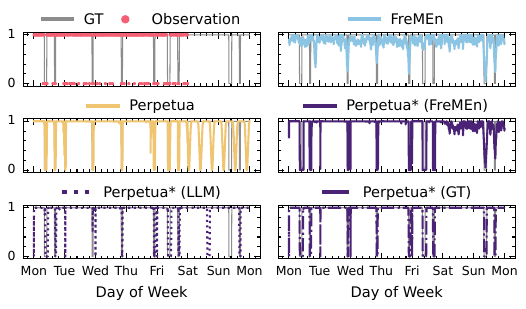}
    \vspace{-20px}
    \caption{\textbf{Persistence Belief:} 
    As shown, FreMEn struggles to adapt to distribution shifts, while Perpetua reverts to a cyclic pattern in the absence of data. \coolPerpetua achieves the best tradeoff between adaptation and prediction quality regardless of the choice of switching prior.
    \vspace{-2em}
    }
    \label{fig:pred_odd}
\end{figure}

We evaluate \coolPerpetua and \coolName in two stages. First, in Sec.~\ref{sec:eval_perpetua}, we benchmark \coolPerpetua against established baselines, assessing long-term prediction and adaptation using a semi-static simulator. Second, in Sec.~\ref{sec:eval_pg}, we evaluate \coolName on semi-static object navigation tasks~\cite{jiang2025dualmap}. We validate our approach across multiple ProcTHOR scenes~\cite{deitke2022️} and demonstrate its effectiveness through real-world deployment in a laboratory with semi-static objects. All time-dependent parameters are reported in seconds.

\subsection{Evaluation: \coolPerpetua}
\label{sec:eval_perpetua}

\subsubsection{Experimental Setup}
\label{sec:details_perpetua_star}

To evaluate \coolPerpetua, we report the \ac{MAE} between the predicted belief and ground truth state, the \ac{BAC} to account for class imbalance (e.g., mostly present or absent features), and the F1 score~\cite{saavedra2025perpetua}. For \ac{BAC} and F1, we threshold the predicted belief at $0.5$; MAE uses the continuous belief.\looseness-1 

Unless stated otherwise, the prior $p_l$ in~\eqref{eq:perpetua_model} is modelled as a mixture of up to five log-normal distributions, with the number of components selected by the AIC as in~\cite{saavedra2025perpetua}. We set the forgetting factor $\gamma = 0.99$ (maintaining the last $250$ observations in memory) and decay rate $\alpha_0 = \num{2.55e-4}$, which corresponds to deferring to the switching prior after $5$ hours. We evaluate \coolPerpetua  with three variants of switching priors $f(t)$ (see \eqref{eq:priors}): (1) a FreMEn prior that computes 1000 Fourier coefficients and selects the optimal subset using a held-out validation set; (2) an LLM prior (\texttt{gemini-2.5-pro}~\cite{comanici2025gemini}; prompt in App.~\ref{sec:weekly_llm_prompt}); and (3) an oracle prior that uses step functions matching the ground truth training set dynamics.\looseness-1 

\subsubsection{Simulation Environment}
\label{sec:procthor_sim}
We use the semi-static simulation environment from Perpetua~\cite{saavedra2025perpetua}, which consists of a robot observing eight objects (four static, four semi-static) that follow a weekly schedule of appearance and disappearance. We collect a train set of one month of data and evaluate on a subsequent one-week test set, applying an observation noise of $P_M = P_F = 0.1$ to both sets. For the train set, patterns are stable during weekdays but exhibit variations during weekends. To demonstrate adaptation capabilities, we introduce a ``long weekend'' scenario in the test set, where two weekdays (Monday and Friday) adopt the weekend behavior (see Fig.~\ref{fig:test_schedule} in App.~\ref{sec:perpetua_data}). This tests the method's ability to adapt when schedules suddenly change.\looseness-1

\begin{table}[t]
\centering
\renewcommand{\arraystretch}{1.05}
\setlength{\tabcolsep}{3.25pt}
\caption{Persistence Estimation Results} 
\begin{tabular}{lccc}
\toprule
Method & MAE $\downarrow$ & B-Acc $\uparrow$ & F1 $\uparrow$ \\
\midrule
Perpetua~\cite{saavedra2025perpetua}  & $0.066 \pm 0.003$ & $0.662 \pm 0.005$ & $0.930 \pm 0.008$  \\
FreMEn~\cite{krajnik2017fremen}  & $0.151 \pm 0.001$ & $0.659 \pm 0.002$ & $0.936 \pm 0.001$  \\
\coolPerpetua (FM) & $\mb{0.042} \pm 0.001$ & $\mb{0.690} \pm 0.010$ & $\mb{0.977} \pm 0.001$  \\ 
\coolPerpetua (LLM) & $0.044 \pm 0.003$ & $0.660 \pm 0.005$ & $0.966 \pm 0.004$  \\ 
\midrule
\rowcolor{gray!20}
\coolPerpetua (GT) & $0.024 \pm 0.002$ & $0.702 \pm 0.012$ & $0.983 \pm 0.001$  \\ 
\bottomrule
\end{tabular}
\label{tab:sim_room_odd}
\vspace{0.5em}
\begin{minipage}{0.95\columnwidth}
    \footnotesize
    \textcolor{black!80}{We ablate \coolPerpetua with different switching priors: FreMEn (FM), an \acs{LLM}, and ground truth (GT). Results averaged over five random seeds.}
\end{minipage}
\vspace{-1.5em}
\end{table}

Table~\ref{tab:sim_room_odd} presents the quantitative results of this evaluation. Overall, \coolPerpetua outperforms Perpetua and FreMEn across all three prior choices, except in \ac{BAC} under the LLM prior. This is expected, since \ac{BAC} averages recall over both classes, so a prior biased toward one state is penalized heavily on the other, even when overall errors are low. Nevertheless, the adaptation capabilities of \coolPerpetua allow it to surpass the baselines on MAE and F1, showing that \coolPerpetua benefits even from a coarse prior. Notably, Perpetua struggles to predict long-term dynamics in the absence of observations (see Saturday and Sunday in Fig.~\ref{fig:pred_odd}), defaulting to cyclic pattern predictions at best. Meanwhile, FreMEn's lack of adaptation capabilities leads to degraded predictions when the test set's dynamics deviate from the training set's (see Monday and Friday in Fig.~\ref{fig:pred_odd}). In contrast, \coolPerpetua offers accurate long-term predictions even in the absence of observations. Lastly, \coolPerpetua has a constant prediction time of $0.1$ms, while Perpetua scales linearly with simulation steps, becoming significantly slower as the switching horizon grows (see Fig.~\ref{fig:complexity}).\looseness-1

\subsection{Evaluation: \coolName}
\label{sec:eval_pg}

\subsubsection{Experimental Setup}

We conduct three distinct assessments: (1) \textit{Predictive Navigation (Simulation)}: 250 predictive closed-vocabulary queries across 10 ProcTHOR \cite{deitke2022️} environments (shown in Fig.~\ref{fig:procthor_environments}); (2) \textit{Adaptive Navigation (Simulation)}: 125 queries across 5 ProcTHOR scenes where test-time object dynamics differ from training (simulated scenes have at most 25 receptacles and up to 20 objects); and (3) \textit{Real-World Experiment}: 100 open-vocabulary queries in a laboratory with three rooms (Fig.~\ref{fig:real_pcd}) containing a total of 28 receptacles and 21 objects, replicating the adaptive navigation setting. Additional simulation and real-world details are given in App.~\ref{app:sim} and App.~\ref{app:real_details}, respectively.

For each query, we randomly sample: (1) a future time point up to \emph{one week} ahead from the last map update, (2) a target object (either static or dynamic, where we differ from~\cite{jiang2025dualmap} by classifying objects that move within their receptacle as static), and (3) a starting robot pose. The robot has up to two attempts to locate and retrieve the target, and a query is successful if the object is visible within 1 meter of the robot's final position~\cite{savva2019habitat}. For the real-world experiment shown in Fig.~\ref{fig:doors}, the initial position of the robot is the same across all queries.
Navigation is based on A* over a map created with RTAB-Map \cite{labbe2019rtab}, and we use an Agilex Ranger Mini 2.0 robot equipped with an Intel RealSense D435 RGB-D camera.

Object-receptacle assignments follow a predefined schedule (in simulation, generated by the hierarchical model in App.~\ref{app:sim_schedule}), executed manually in the real world. The environment updates hourly in simulation and every two hours in the real world; at each update, semi-static objects may move, vanish, or reappear, and the robot collects mapping data between updates. We generate four weeks of data per simulated environment (split 2:1:1 for training, validation, and testing) and three weeks for the real-world experiments.

We report \textit{\ac{SR}} and \textit{\ac{SPL}}. For \textit{absent} objects, we additionally report the correct \textit{\ac{AR}} and the \textit{\ac{DIST}} before termination. Standard deviations are reported across simulated scenes and 95\% confidence intervals in the real world.\looseness-1

\begin{figure}
    \centering
    \includegraphics[clip, width=\linewidth]{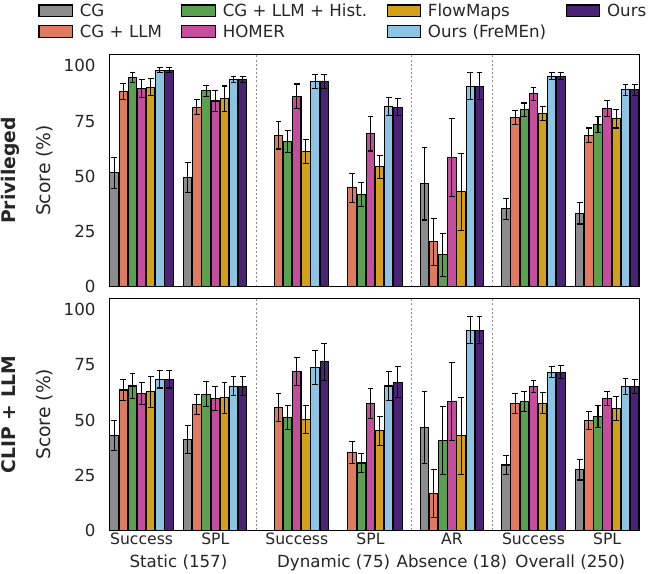}
    \vspace{-2em}
    \caption{\textbf{Predictive Navigation:} \coolName with \coolPerpetua or FreMEn estimators outperforms baselines under both privileged and noisy perception. Query counts in parentheses and error bars denote one standard deviation. \looseness-1\vspace{-2em}}
    \label{fig:main_batplot}
\end{figure}%

\subsubsection{Baselines}

We compare \coolName against seven baselines:
(i) \textbf{ConceptGraphs (CG)}~\cite{gu2024conceptgraphs}, an open-vocabulary 3D scene graph capturing spatio-semantic relations where object locations are retrieved from the last map state;
(ii) \textbf{CG + LLM}, which augments CG with an \acs{LLM} to reason about object dynamics given the time elapsed since the last map update;
(iii) \textbf{CG + LLM + History}, which, inspired by~\cite{anwar2025remembr}, extends CG + \acs{LLM} with a history of the last 100 timestamped receptacle observations;
(iv) \textbf{DualMap}~\cite{jiang2025dualmap}, an open-vocabulary scene graph maintaining dual global-local maps to handle semi-static scenes; 
(v) \textbf{PredictiveGraphs (FreMEn)}, an ablation of our method replacing our estimator, \coolPerpetua, with FreMEn~\cite{krajnik2017fremen} as the switching prior; (vi) \textbf{HOMER}~\cite{homer}, a temporal link prediction method based on a graph neural network (GNN); and (vii) \textbf{FlowMaps}~\cite{flowmaps}, a learning method based on flow matching that can predict patterns across multiple different scenes.\looseness-1

\subsubsection{Training Details}
We train \coolPerpetua following the procedure in Sec.~\ref{sec:details_perpetua_star} with decay rate $\alpha_0 = \num{2.56e-3}$ (switching to the prior after 30 minutes). We defer to the switching prior faster than in Sec.~\ref{sec:eval_perpetua}, whose long-weekend dynamics deviate from training, making the prior less reliable. Our base scene graph is ConceptGraphs~\cite{gu2024conceptgraphs} and we classify objects as present or absent based on a threshold of $\delta=0.2$. All methods use the same association function based on CLIP~\cite{radford2021learning} embeddings with an LLM verifier (\texttt{gpt-5-mini}).

\subsubsection{Simulation Results}
\label{sec:sim_results}

Figure~\ref{fig:main_batplot} shows the predictive navigation results. To isolate the impact of perception errors, we also report privileged results (using ProcTHOR's~\cite{deitke2022️} ground-truth segmentation, still subject to occlusion). \coolName achieves the best performance across all query types with both FreMEn and \coolPerpetua\ as estimators. Notably, while the LLM-based baselines are competitive, their performance degrades as they predict typical object relocation patterns rather than learning environment-specific dynamics. See Table~\ref{tab:predictive_sim} for tabulated results, Fig.~\ref{fig:qualitative_sim} for qualitative examples, and Fig.~\ref{fig:failure_sankey} for a Sankey diagram analyzing failures.\looseness-1

\begin{figure}[t]
    \centering
    \includegraphics[width=\linewidth]
    {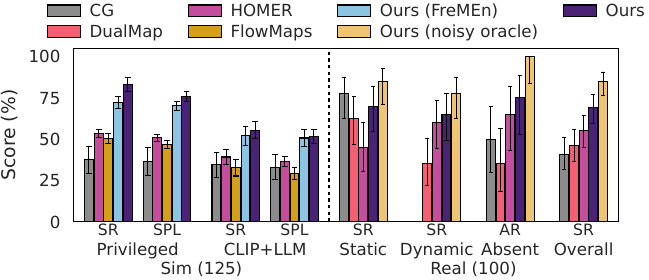}
   \vspace{-20px} \caption{\textbf{Adaptation in Simulation (left; error bars: std) and Real-World Navigation (right; error bars: 95\% CI).} The adaptive capabilities of \coolPerpetua enable better performance even under distributional shifts in environment dynamics. \coolName enables robust dynamic navigation even when trained on a noised GT schedule, or under real-world perception errors (table versions of this figure and Fig.~\ref{fig:main_batplot} in Tables~\ref{tab:adaptive_sim_full} (left), \ref{tab:adaptive_real_world} (right), and~\ref{tab:predictive_sim}).\looseness-1}\vspace{-1.5em}
    \label{fig:barplots}
\end{figure}

The adaptive navigation results are shown in Fig.~\ref{fig:barplots} (left) and Table~\ref{tab:adaptive_sim_full}, with an example of the train/test shift in object dynamics in Fig.~\ref{fig:procthor_schedule}. We ablate \coolName using a fully-predictive estimator (FreMEn) against our adaptive \coolPerpetua. Under privileged perception, \coolName with \coolPerpetua achieves better navigation performance due to its ability to adapt to changing underlying dynamics. Even with perception errors degrading overall performance, \coolName with \coolPerpetua continues to outperform the FreMEn variant. We also note that, as perception pipelines improve, we expect \coolName's performance to improve further. 
Across simulation experiments, both learned methods~\cite{homer, flowmaps} underperform \coolName. We attribute this to their data requirements: unlike \coolName, which fits from little data and can also adapt to new patterns online, they must be trained offline on large corpora of environment dynamics before deployment.\looseness-1

\subsubsection{Real-World Results}
\label{sec:real_world_results}

\begin{figure}
    \centering
    \includegraphics[width=\linewidth]{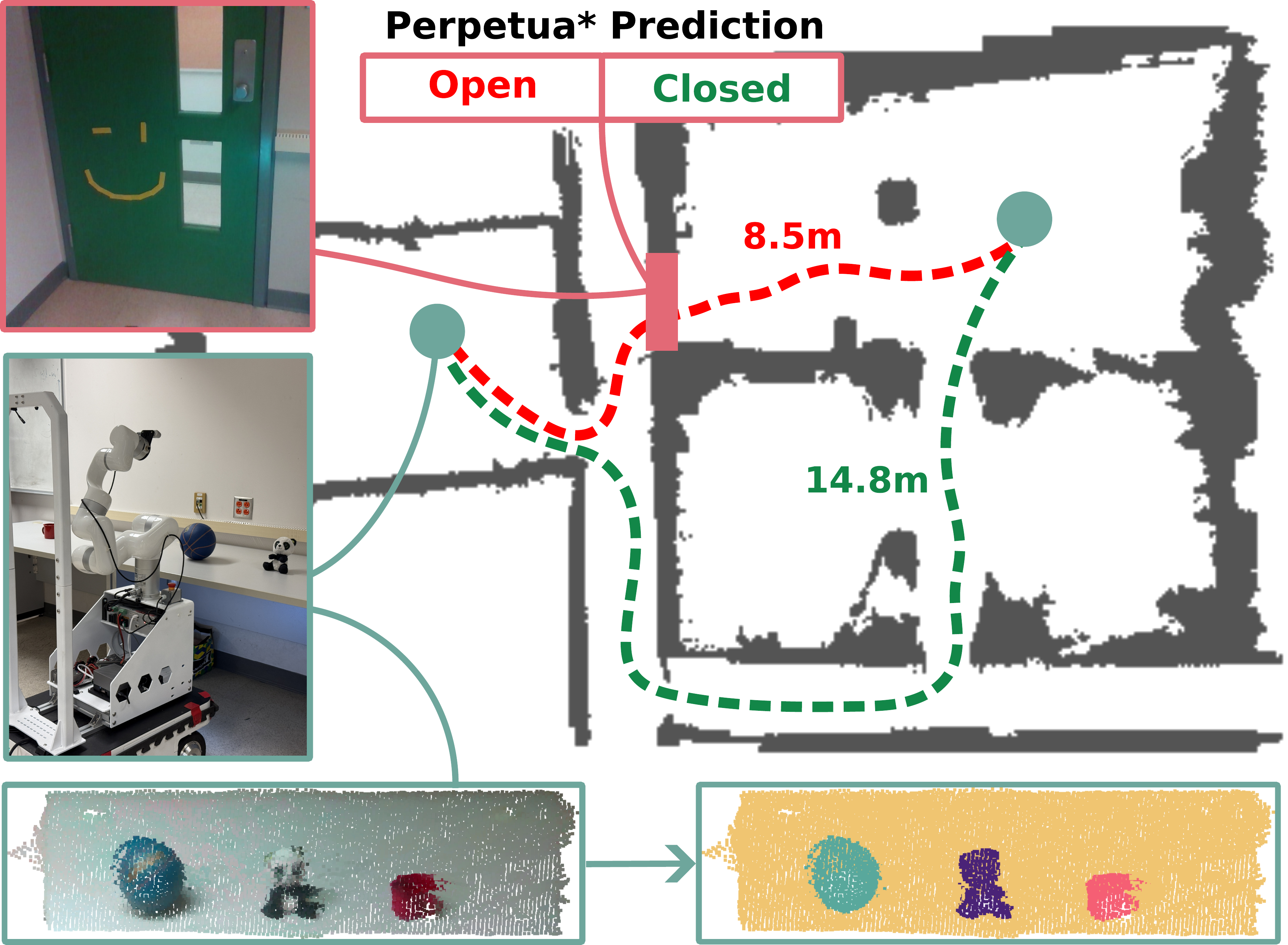}
    \caption{
    \textbf{Predictive Navigation in the Real World:} \coolName uses its predictive capabilities to anticipate that the shortest path to its goal is blocked ({\color{pink}low probability}). This allows it to preemptively adapt its navigation plan, choosing a path that is longer but feasible ({\color{teal}high probability}), and successfully reaching the target without visiting unnecessary locations.\looseness-1\vspace{-2em}}
    \label{fig:doors}
\end{figure}

We showcase how \coolName's temporal prediction enables effective dynamic navigation under real-world perception noise. We compare two training regimes: training on a noised version (10\% noise) of the ground truth schedule versus training on data labelled using the procedure described in Sec.~\ref{sec:pg_method}. As shown in Fig.~\ref{fig:barplots} (right), \coolName robustly estimates semi-static changes even when the scene graph is outdated and perception is noisy, surpassing both DualMap~\cite{jiang2025dualmap}, a method designed for semi-static environments but lacking predictive capabilities, and HOMER~\cite{homer}, a GNN-based predictive method designed for semi-static changes. FlowMaps is not evaluated in the real world, as it requires multi-environment training data unavailable in this setting. Fig.~\ref{fig:doors} further illustrates the value of our approach: \coolName anticipates a blocked shortest path and preemptively selects a longer but viable alternative towards its target, avoiding unnecessary exploration. The full sequence of Fig.~\ref{fig:doors} and an additional replanning example are shown in Figs.~\ref{fig:qualitative_doors} and~\ref{fig:qualitative_replanning}.\looseness-1

Finally, the memory footprint of our method is negligible (Table~\ref{tab:memory_footprint}):  $0.047$~KB per mixture component, or $28$~KB per edge for a model with five mixture components and a 1,000-coefficient FreMEn prior. Thus, a 1{,}000-edge graph adds \textbf{28~MB} atop a base scene graph that can itself reach several~GB.

\section{Conclusion} 
\label{sec:conclusion}

In this work, we present \coolName, a new 3D scene graph representation that is capable of admitting tempo-spatio-semantic queries. Underpinning the representation is a new persistence filter that we propose called \coolPerpetua. Our filter uses tempered Bayesian model averaging to combine emergence and persistence filters, capturing both the appearance and disappearance of objects over time. The combination of scene graphs and \coolPerpetua (\coolName) is powerful in that it allows us to predict what the semantic scene graph will be in the future and then directly reason over it. Future generalist robots will require these types of reasoning capabilities to enable seamless collaboration with humans and the ability to achieve complex tasks. Even today, surveillance applications such as factory monitoring can already benefit from our approach by providing additional temporal awareness.  Future work will relax our receptacle assumptions, enabling \coolName to track receptacles that themselves move and to reason about object-receptacle pairs it has never observed.\looseness-1

\section*{Acknowledgments}
We acknowledge the support of the Natural Sciences and Engineering Research Council of Canada (NSERC) [PGS D Scholarships for Charlie Gauthier and Miguel Saavedra-Ruiz]. This work was also supported by CIFAR under the Canada CIFAR AI Chairs program and by the NSERC under the Discovery Grants Program, and in part by compute resources provided by Mila (mila.quebec).

\bibliographystyle{IEEEtran}
\bibliography{IEEEabrv, references}

\newpage

\clearpage

\newpage

\appendices

\renewcommand{\topfraction}{0.9}
\renewcommand{\bottomfraction}{0.8}
\renewcommand{\textfraction}{0.07}
\renewcommand{\floatpagefraction}{0.8}
\renewcommand{\dbltopfraction}{0.9}
\renewcommand{\dblfloatpagefraction}{0.8}
\setcounter{topnumber}{3}
\setcounter{bottomnumber}{2}
\setcounter{totalnumber}{5}
\setcounter{dbltopnumber}{2}

\renewcommand\thesubsection{\thesection.\arabic{subsection}}
\makeatletter
\gdef\thesubsection{\thesection.\arabic{subsection}}
\makeatother
\renewcommand{\thesubsection}{\thesection.\arabic{subsection}}
\renewcommand{\thesubsectiondis}{\thesection.\arabic{subsection}}

\section{Supplementary Details: \coolPerpetua}

\subsection{\coolPerpetua Algorithm}
\FloatBarrier
\label{app:algorithm}

In Algorithm~\ref{alg:mixture_filter}, we present the update and prediction routines for a single receptacle in \coolPerpetua, omitting the receptacle index $k$ for clarity. We also use the shorthand notation $\rho_{N+1} \triangleq P_F^{y_{t_{N+1}}} (1-P_F)^{1-y_{t_{N+1}}}$, $ H_{l,N} \triangleq H(\mathcal{Y}_{1:N} \given c_l, m)$, and $\Delta F_{l,N} \triangleq F_l(t_N) - F_l(t_{N-1})$ where we made explicit the dependence on the mixture component $l$ and model $m$.  During the update step, we compute the likelihood, conditional evidence, and posterior weights for both the persistence and emergence models. In the prediction step, we generate the \coolPerpetua prediction at an arbitrary query time using our proposed tempered Bayesian model averaging.

\begin{algorithm} 
\caption{Predict and Update Functions for \coolPerpetua\ (Single Receptacle) } 
\label{alg:mixture_filter}

\SetKwInput{KwParam}{Parameters}
\SetKwInput{KwInUpdate}{Update Input}
\SetKwInput{KwInPredict}{Predict Input}
\SetInd{0.5em}{0.5em}
\SetKwFunction{FUpdate}{Update}
\SetKwFunction{FPredict}{Predict}

\KwParam{Models $m \in \{\Mper, \Memg\}$, Detector params $P_M, P_F$, Decay rate $\alpha_0$, Forgetting factor $\gamma$ for receptacle $k$}
\SetKwComment{Comment}{$\triangleright$\ }{}

\vspace{0.025cm}

\Comment{\small{Updates the state of each model; for $\Memg$ substitute $(P_M, P_F) \leftarrow (1-P_F, 1-P_M)$}}
\KwInUpdate{Observation tuple ($y_{t_{N+1}}, t_{N+1}$)}
\SetKwProg{Fn}{Function}{:}{}
\Fn{\FUpdate{$y_{t_{N+1}}, t_{N+1}$}}{
    \For{$m \in \{\Mper, \Memg\}$}{
        Update likelihood $\ell_{N+1}$ in~\eqref{eq:perpetua_likelihood} with $\gamma$:\hfill\break
        $\ell_{N+1} \leftarrow \ell_{N}^{\gamma} \, P_M^{1-y_{t_{N+1}}} (1-P_M)^{y_{t_{N+1}}}$\;
        \For{$l \gets 1$ \KwTo $L$}{
            Update accumulator $H_{l,N+1}$ in~\eqref{eq:perpetua_evidence} with $\gamma$:\hfill\break
$H_{l,N+1} \leftarrow \rho_{N+1} ( H_{l,N}^{\gamma} + \ell_{N}^{\gamma} \Delta F_{l,N+1})$\;
            {\small $p(\mathcal{Y}_{1:{N+1}} \hspace{-0.5mm }\given \hspace{-0.5mm} c_l, m) \hspace{-0.5mm} \leftarrow H_{l,N+1} + \ell_{N+1} [1 - F_l(t_{N+1})]$}\;
        }
        Update posterior weights $w_l$ via \eqref{eq:perpetua_weights}\;
    }
        $N \leftarrow (N + 1)$\;
}

\vspace{0.025cm}

\Comment{\small{Predicts persistence state at time $t$}}
\KwInPredict{Query time $t$ for receptacle $k$}
\SetKwProg{Fn}{Function}{:}{}
\Fn{\FPredict{$t$}}{
    \For{$m \in \{\Mper, \Memg\}$}{
    Evaluate predictions $p(x_t \given c_{l^*}, \mathcal{Y}_{1:N}, m)$ via \eqref{eq:perpetua_mode}\;
    }
    $\alpha(t) \leftarrow \exp(-\alpha_0 (t - t_N))$\;
    Compute model posterior $p(\mathcal{M} \mid \mathcal{Y}_{1:N})$ via \eqref{eq:perpetua_model_selection}\;
    Compute final prediction $p(x_t \given \mathcal{Y}_{1:N})$ via \eqref{eq:perpetua_final_pred}\;
    \KwRet $p(x_t \mid \mathcal{Y}_{1:N})$\;
}
\end{algorithm}

\section{Supplementary \& Implementation Details}
\label{app:impl_details}

\subsection{LLM Tools}
\label{app:tools}

In this section, we provide the formal specifications for the three core tools that enable embodied planning in \coolName: \emph{semantic search} (Tool~\ref{tools:tool1}), \emph{location prediction} (Tool~\ref{tools:tool2}), and \emph{active navigation} (Tool~\ref{tools:tool3}). For each tool, we define the required inputs, the expected outputs, and the underlying logic used to interface with \coolName.

\begin{toolBox}{Semantic Object Search}{tools:tool1}
\textbf{Input:} \texttt{query\_text} (string) \\
\textbf{Output:} List of $\langle$ \texttt{object\_id}: string, \texttt{score}: float $\rangle$ or \texttt{None}\\[0.5em]
\textbf{Description:} Performs an open-vocabulary search over \coolName using CLIP. It compares the embedding of the \texttt{query\_text} against the visual embeddings of all mapped objects, returning the \texttt{object\_ids} with the highest cosine similarities.
\end{toolBox}

\begin{toolBox}{Predict Object Receptacle}{tools:tool2}
\textbf{Input:} \texttt{object\_id} (string), \texttt{query\_time} (float) \\
\textbf{Output:} Most likely receptacle name (string), probability distribution over receptacles (list of floats) \\[0.5em]
\textbf{Description:} Queries the temporal object map (\coolName) using \coolPerpetua to predict the location of the target object at the specified \texttt{query\_time}. The tool applies a confidence threshold (default $\delta = 0.2$) to filter receptacles. If no receptacle exceeds the threshold, the object is reported as absent.
\end{toolBox}

\begin{toolBox}{Navigate to Receptacle}{tools:tool3}
\textbf{Input:} \texttt{receptacle}\_\texttt{id} (string), \texttt{object}\_\texttt{id} (string) \\
\textbf{Output:} Success status (boolean), final observation \\[0.5em]
\textbf{Description:} Navigates to the \texttt{receptacle\_id} while actively searching for the target \texttt{object\_id}. During navigation, the agent continuously: (1) executes path planning toward the receptacle, (2) updates \coolName with RGBD observations, and (3) checks whether the target object is in view. Navigation terminates when either the object is found or the target receptacle is reached.
\end{toolBox}

\subsection{\acs{LLM} Prompt - Switching Prior} 
\label{sec:weekly_llm_prompt}

We present the full \acs{LLM} prompt used to query the LLM-based switching prior described in Sec.~\ref{sec:details_perpetua_star}. We employ a few-shot prompting strategy to leverage the semantic knowledge of \texttt{gemini-2.5-pro}, enabling it to infer probable weekly schedules for objects based on their household context and potential usage patterns. The prompt is shown in Prompt~\ref{prompt:llm_prior}.

\begin{promptBox}{LLM Prior for \coolPerpetua}{prompt:llm_prior}
Consider a household environment composed of two parents and two children. The environment contains various objects located in different places, where objects can appear when needed for an activity (e.g., plates for eating) or disappear when taken by a member (e.g., keys/bags when leaving). During weekdays, the parents typically leave for work in the morning and return in the evening (07:30 - 18:00), while the children attend school during the day (08:00 - 17:00). On weekends, the family tends to stay home more often, engaging in activities such as cooking, cleaning, and leisure time together. \\

\textbf{Task}: Generate a weekly schedule (Mon-Sun) for the specific object at the specific location. \\

\textbf{Output Rules}:
\begin{itemize}
    \item Output ONLY a Python list of strings.
    \item Format: ['Day: HH:MM - Day: HH:MM', 'Day: HH:MM - Day: HH:MM', ...]
    \item Use 3-letter day abbreviations (Mon, Tue, Wed, Thu, Fri, Sat, Sun).
\end{itemize}

\textbf{Example 1:}\\
Input:\\
\hspace*{1.5em} \textit{Object}: Plates \\
\hspace*{1.5em} \textit{Location}: On Dinner Table \\
\hspace*{1.5em} \textit{Type}: Active Location (appear when used) \\
Output: \\
\hspace*{1.5em} \textit{Schedule}: ['Mon: 19:00 - Mon: 19:30', 'Tue: 19:00 - Tue: 19:30', ... , 'Sun: 18:00 - Sun: 18:30']

\textbf{Current task}: \\
Input: \\ 
\hspace*{1.5em} \textit{Object}: \{object\_name\} \\
\hspace*{1.5em} \textit{Location}: \{object\_location\} \\
\hspace*{1.5em} \textit{Type}: \{object\_type\} \\
Output: \\
\hspace*{1.5em} \textit{Schedule}: [...]
\end{promptBox}

\subsection{\coolPerpetua Implementation Details}
\label{app:perpetua_impl}

In this section, we detail the parameters and modeling choices for \coolPerpetua.

\paragraph{Hyperparameters} The prior $p_l$ in~\eqref{eq:perpetua_model} is modeled as a mixture of log-normal distributions. We set the forgetting factor $\gamma = 0.99$, which maintains an effective memory of approximately $250$ observations. The decay rate is set to $\alpha_0 = \num{2.55e-4}$, causing the model to revert almost entirely to the prior after approximately $5$ hours.

\paragraph{Model Selection} Following the configuration in Perpetua~\cite{saavedra2025perpetua}, we search for the optimal number of mixture components (up to a maximum of $5$) for both persistence and emergence filters using the \ac{AIC}~\cite{akaike1974}. The observation noise parameters ($P_M, P_F$) are estimated directly from dataset statistics.

\paragraph{Switching Priors} We evaluate \coolPerpetua with three variants of the switching prior $f(t)$:

\begin{enumerate}\item 

\textbf{FreMEn Prior:} Based on~\eqref{eq:priors}, this variant computes 1000 Fourier coefficients and selects the optimal subset via a held-out validation set of one week.

\item \textbf{LLM Prior:} Generated via \texttt{gemini-2.5-pro}~\cite{comanici2025gemini} using the formulation in~\eqref{eq:priors}. The full prompt is shown in App.~\ref{sec:weekly_llm_prompt}.

\item \textbf{Oracle Prior:} Uses step functions that perfectly match the ground truth dynamics of the training set.

\end{enumerate}

\subsection{Schedule}
\label{sec:perpetua_data}

\begin{figure}[!tbp]
    \centering
    \includegraphics[width=\columnwidth]{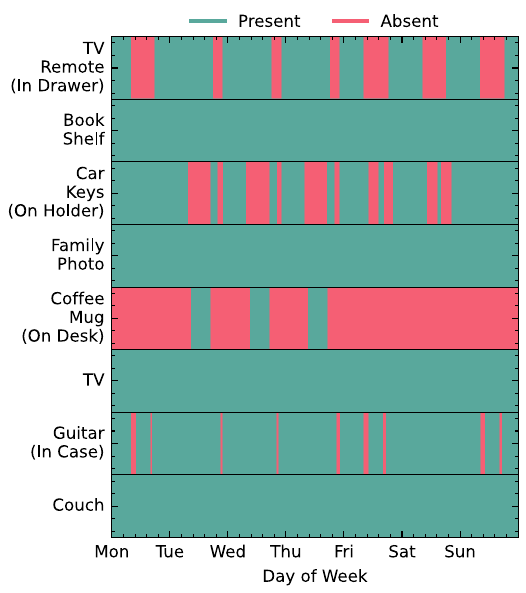}
    \caption{\textbf{Weekly Test Schedule:} One-week snapshot of the test set with out-of-distribution dynamics for the different objects used in the \coolPerpetua evaluation presented in Sec.~\ref{sec:eval_perpetua}. Note how Friday and Monday follow the same dynamics as the weekend, emulating a ``long weekend''.}
    \label{fig:test_schedule}
\end{figure}

While the training set used to train \coolPerpetua in Sec.~\ref{sec:eval_perpetua} reflects a standard weekly routine with distinct weekday and weekend patterns, the test set introduces a distributional shift to evaluate adaptation.  Specifically, we construct a ``long weekend'' scenario where Monday and Friday exhibit weekend-like dynamics. This modification forces the model to adjust to sudden schedule changes, serving as a stress test for real-time adaptation. A snapshot of these shifted dynamics is provided in Fig.~\ref{fig:test_schedule}.

\subsection{Hardware}
\label{app:real_impl_details}

All simulation experiments were conducted on an institutional cluster equipped with Nvidia RTX 8000 and L40S GPUs. The Agilex Ranger Mini 2.0 was equipped with an Nvidia Jetson Orin for onboard control, but model inference was performed offboard, on an Nvidia RTX 2080. For both the LLM-based navigation agent and the LLM-based predictive baselines we use \texttt{gpt-5.2}.

\section{Supplementary Results: \coolPerpetua}

\subsection{Complexity Analysis and Illustrative Example: \coolPerpetua}

\label{app:complexityanalysis}

\begin{figure}[!tbp]
    \centering
    \includegraphics[width=\linewidth]{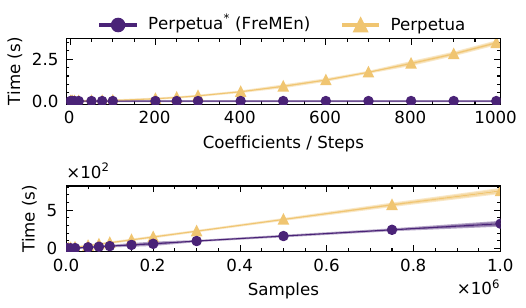}
    \caption{\textbf{Computational Complexity:} prediction (top) and update (bottom) steps for {\color{purple}\coolPerpetua with a FreMEn prior} versus {\color{orange}Perpetua}. Top: prediction time as a function of Fourier coefficients in \coolPerpetua and simulation steps in Perpetua. Bottom: update time with up to $1$ million samples. Results are averaged over five random seeds. \coolPerpetua exhibits faster computation in both cases by replacing Perpetua's state machine with our tempered Bayesian model averaging.}
    \label{fig:complexity}
\end{figure}

In Fig.~\ref{fig:complexity}, we report the results of a complexity analysis of \coolPerpetua versus Perpetua. \coolPerpetua is more computationally efficient at both prediction and belief updates due to its switching mechanism based on tempered Bayesian model averaging, which avoids Perpetua's reliance on expensive simulation steps. 

\begin{figure}[!tbp]
    \centering
    \includegraphics[width=\linewidth]{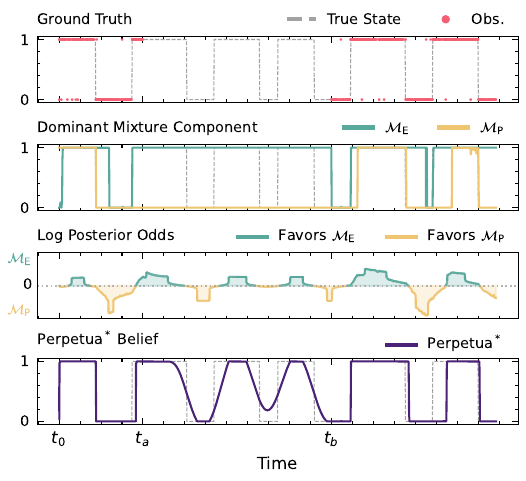}
    \caption{\textbf{Illustrative Example \coolPerpetua}: An object undergoes semi-static changes at varying rates (top row). We maintain a mixture of persistence and emergence filters (each comprising three components in this example) that \emph{adapt} their belief as new observations arrive. The second row shows the dominant component of the persistence and emergence filters (see \eqref{eq:perpetua_mode}) at each time step. As time progresses, our tempered Bayesian model averaging (see \eqref{eq:perpetua_model_selection}) identifies the mixture that best explains the observed data (third row). Notably, between times $t_a$ and $t_b$, where no observations are available, the log posterior odds keep evolving due to our switching prior, whereas the dominant mixture component remains frozen. The bottom row shows the final \coolPerpetua\ prediction, obtained by mixing the output of both mixture models at each time step, as shown in \eqref{eq:perpetua_final_pred}. By leveraging the switching prior (cf.~Fig.~\ref{fig:model_selection} and Fig.~\ref{fig:pred_odd}), \coolPerpetua\ maintains accurate predictions even in the absence of observations.\looseness-1}
    \label{fig:fig_example} 
\end{figure}

Table~\ref{tab:memory_footprint} shows the memory footprint of adding extra mixture components or edges to the graph. The per-mixture cost is negligible ($0.047$~KB), while the per-edge cost is $28$~KB for a model with five mixture components and a FreMEn prior with a thousand coefficients. New edges are only added when a new object-receptacle pair is observed, so memory scales linearly with the number of discovered relationships rather than with deployment time. For example, in a scene with 250 objects appearing in up to 4 different locations each (1,000 edges), the total overhead is \textbf{28~MB} on top of the base scene graph, which can itself reach several~GB due to stored object point clouds. Overall, the additional memory and compute overhead of \coolPerpetua\ is negligible compared to modern mapping approaches. 
 Furthermore, in Fig.~\ref{fig:fig_example} we present an illustrative example showing how the different components (mixture models, tempered Bayesian model averaging) interact to yield the final \coolPerpetua\ prediction even in the absence of data.\looseness-1

\label{app:complexity}

\begin{table}[t]
\centering
\caption{Memory footprint of \coolPerpetua}
\begin{tabular}{llc}
\toprule
& Quantity & Value \\
\midrule
\multirow{2}{*}{(a) Per mixture component}
& Base cost          & $27.89$ KB  \\
& Cost per component & $0.047$ KB  \\
\midrule
\multirow{2}{*}{(b) Per edge}
& Base cost      & $0.000$ KB  \\
& Cost per edge  & $28.13$ KB \\
\bottomrule
\end{tabular}
\begin{minipage}{0.95\columnwidth}
    \footnotesize
    \vspace{0.5em} 
    \textcolor{black!80}{(a)~Per-edge cost as a function of mixture components. (b)~Total cost as a function of edges using mixture models with five mixture components and a FreMEn prior with 1,000 coefficients.}
\end{minipage}
\vspace{-1em}
\label{tab:memory_footprint}
\end{table}

\subsection{Empirical Check of the Independence Assumption}
\label{app:independence}

The multi-receptacle belief in Sec.~\ref{sec:perpetua_star} factorizes the object-receptacle edges independently and therefore does not enforce mutual exclusivity across receptacles. We adopt this approximation deliberately, as a data-efficiency and scalability trade-off. A categorical location model would require learning a joint distribution over all receptacles, including location transitions as well as the location-specific multimodal temporal patterns that \coolPerpetua currently captures. The number of transition parameters grows quadratically with the number of receptacles, while the available data typically contain few relocation events spread across many receptacle pairs, which makes such a model difficult to estimate from sparse and partial observations. In contrast, the independent formulation updates each edge from local observations, supports newly discovered receptacles without retraining a global model, and scales linearly with the number of observed edges (Table~\ref{tab:memory_footprint}).

The factorization also reflects the distinction between a \emph{semantic class} (e.g., ``mug'', or anything matching an open-vocabulary query) and the \emph{physical objects} that instantiate it. When the perception system cannot tell two objects apart, observations of the semantic class at different receptacles are all valid, and multiple receptacle hypotheses are admissible. A single physical object, however, cannot be in two receptacles at once, which is why the argmax in~\eqref{eq:perpetua_map} produces a single prediction when one is required. The argmax does not discard the remaining edge beliefs: these are retained and updated online, and the navigation agent visits receptacles in decreasing order of belief (Sec.~\ref{sec:pg_method}).

\begin{figure*}[!tp]
    \centering
    \includegraphics[width=0.45\textwidth]{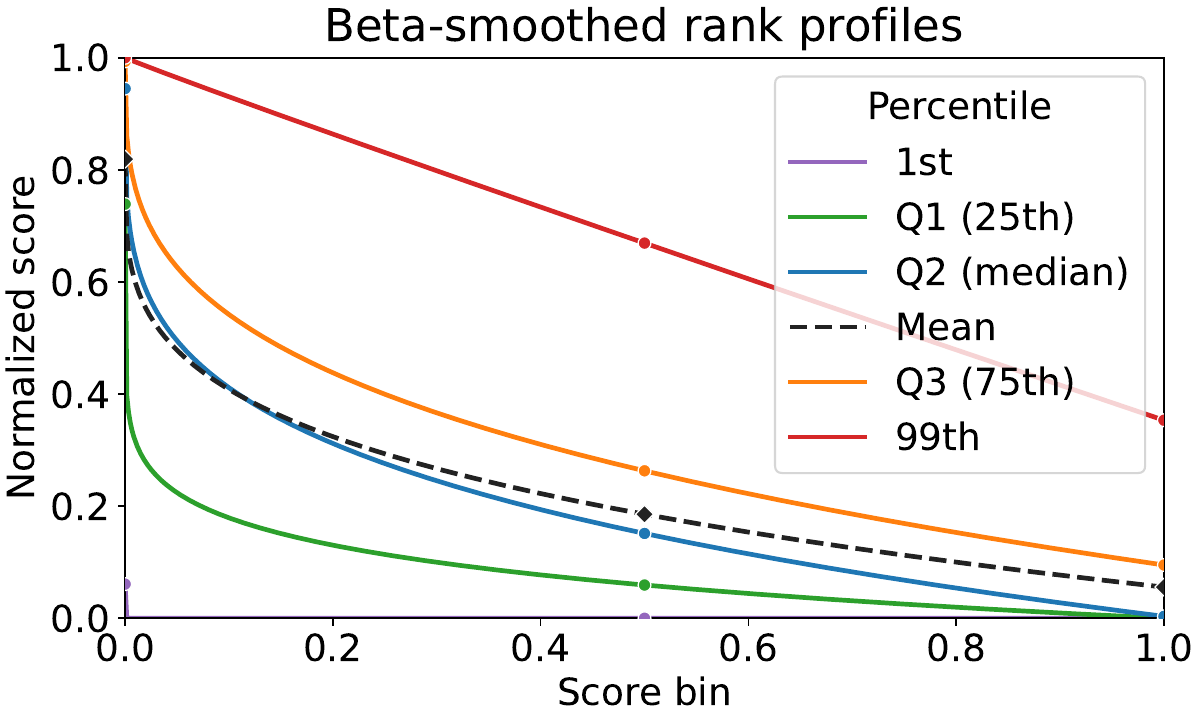}
    \hfill
    \includegraphics[width=0.47\textwidth]{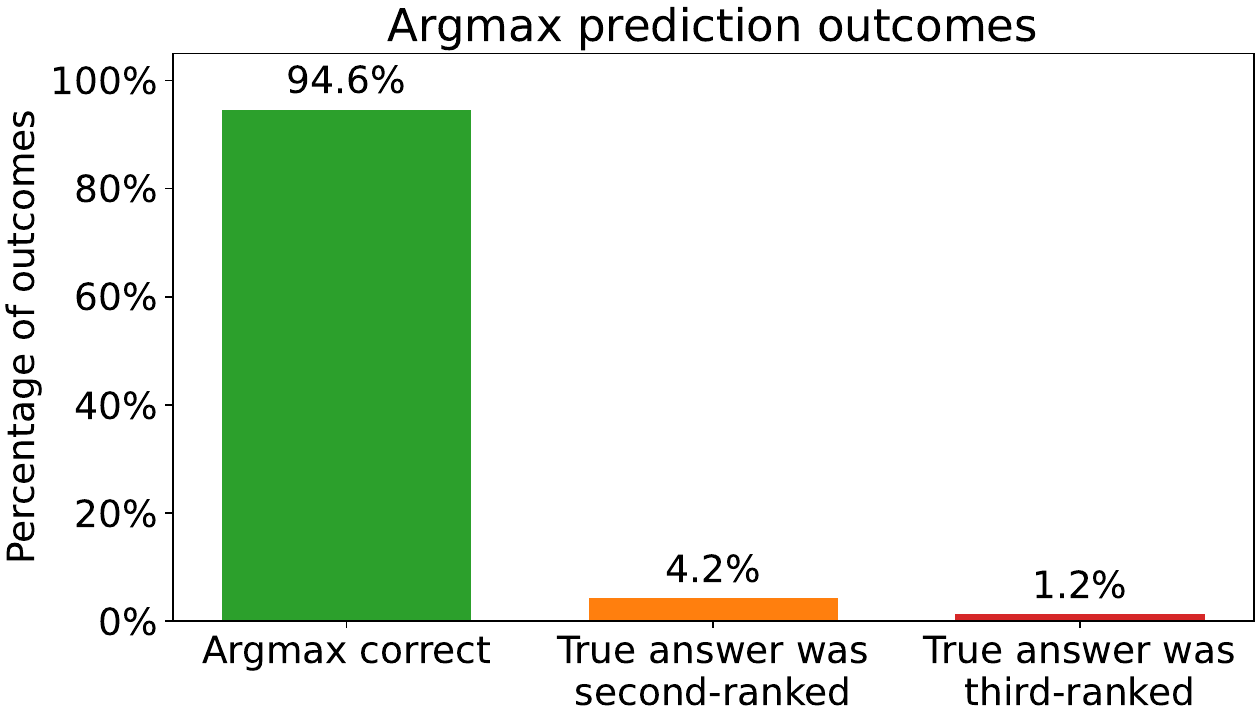}
    \caption{\textbf{Empirical Check of the Independence Assumption:} computed on the queries of the predictive navigation experiment (Fig.~\ref{fig:main_batplot}), using three score bins for visualization. (Left)~Sorted receptacle-score percentile profiles (rank 0 = best): to show how the top-scoring receptacle dominates the score mass, we fit one Beta distribution to each of a suite of percentiles of the score disagreement between receptacles; in all cases, the profile is skewed towards the top-ranked candidate. (Right)~Argmax outcomes on queries where the target object is present: the argmax receptacle is correct 94.6\% of the time (true receptacle ranked second 4.2\% and third 1.2\% of the time).}
    \label{fig:independence}
\end{figure*}

To quantify the effect of this approximation, we analyze the receptacle scores of \coolPerpetua over all queries of the predictive navigation experiment (Fig.~\ref{fig:main_batplot}). As shown in Fig.~\ref{fig:independence} (left), the score mass is concentrated on the top-ranked receptacle, so conflicting high-probability edges are rare. On queries where the target object is present, the argmax receptacle is correct 94.6\% of the time, with the true receptacle ranked second 4.2\% and third 1.2\% of the time (Fig.~\ref{fig:independence}, right).

\subsection{Real-World Evaluation on PKLot}
\label{app:pklot}

To compare \coolPerpetua and Perpetua under real detector noise and naturally occurring dynamics, we evaluate both on the PKLot parking-lot occupancy benchmark~\cite{almeida2015pklot}, also used in the original Perpetua paper~\cite{saavedra2025perpetua}. We use the UFPR04 and UFPR05 lots (28 and 40 parking spaces, respectively) and evaluate \coolPerpetua with FreMEn (FM), LLM, and ground-truth (GT) switching priors. PKLot consists of fixed cameras observing binary parking-space occupancies rather than posed RGB-D observations of object-receptacle relations, so it cannot be used to evaluate \coolName; we use it strictly as a \coolPerpetua-versus-Perpetua comparison.

\begin{figure*}[!tp]
    \centering
    \includegraphics[width=0.9\textwidth]{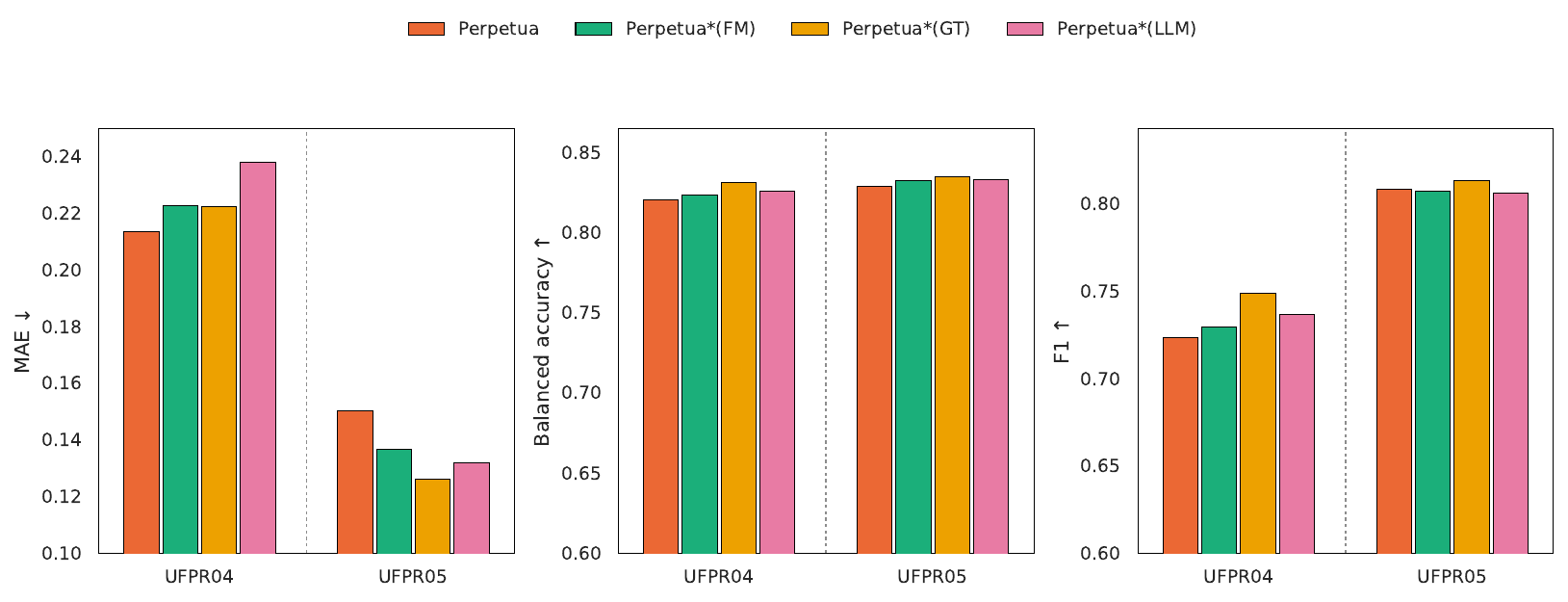}
    \caption{\textbf{PKLot Results:} MAE, balanced accuracy, and F1 on the UFPR04 and UFPR05 lots of PKLot~\cite{almeida2015pklot}, analogous to Table~II of~\cite{saavedra2025perpetua}. \coolPerpetua performs similarly to Perpetua across switching priors.}
    \label{fig:pklot_headline}
\end{figure*}

\begin{figure*}[!tp]
    \centering
    \includegraphics[width=0.8\textwidth]{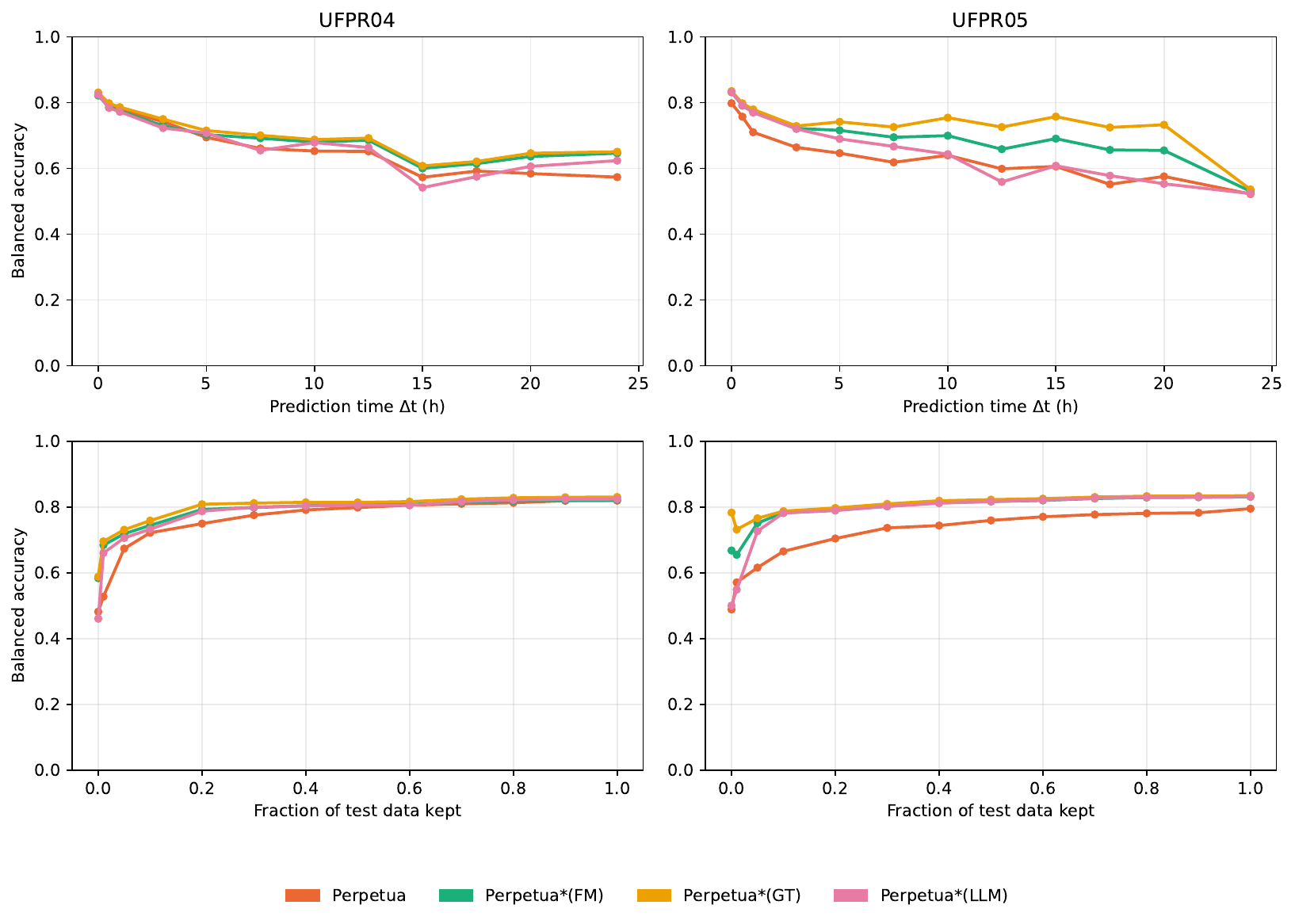}
    \caption{\textbf{PKLot Data Sparsity and Prediction Horizon:} balanced accuracy as a function of the prediction horizon $\Delta t$ (top) and of the fraction of test data kept (bottom), replicating Fig.~4 of~\cite{saavedra2025perpetua}. \coolPerpetua and Perpetua behave similarly in many cases, with \coolPerpetua (FM) holding a slight edge. \coolPerpetua (GT) shows that privileged knowledge injected through the switching prior can lead to large performance improvements.}
    \label{fig:pklot_sparsity}
\end{figure*}

Figure~\ref{fig:pklot_headline} reproduces the experiment in Table~II of~\cite{saavedra2025perpetua}. \coolPerpetua performs on par with or slightly better than Perpetua on most metrics, the exception being MAE on UFPR04, where Perpetua is lower. Figure~\ref{fig:pklot_sparsity} replicates the data-sparsity and prediction-horizon study in Fig.~4 of~\cite{saavedra2025perpetua}. \coolPerpetua and Perpetua behave similarly in many cases, but \coolPerpetua (FM) has a slight edge when observations are missing. \coolPerpetua (GT) illustrates a useful property of our formulation: privileged, human-derived knowledge can easily be injected through the switching prior, which is not possible with Perpetua's switching mechanism, and can lead to large performance improvements. Finally, the main practical advantage of \coolPerpetua over Perpetua is computational (Fig.~\ref{fig:complexity}), making it better suited to large-scale deployments with many receptacles.

\section{Simulation Details: \coolName}
\label{app:sim}

\begin{figure*}[!tp]
    \centering
    \includegraphics[width=\linewidth]{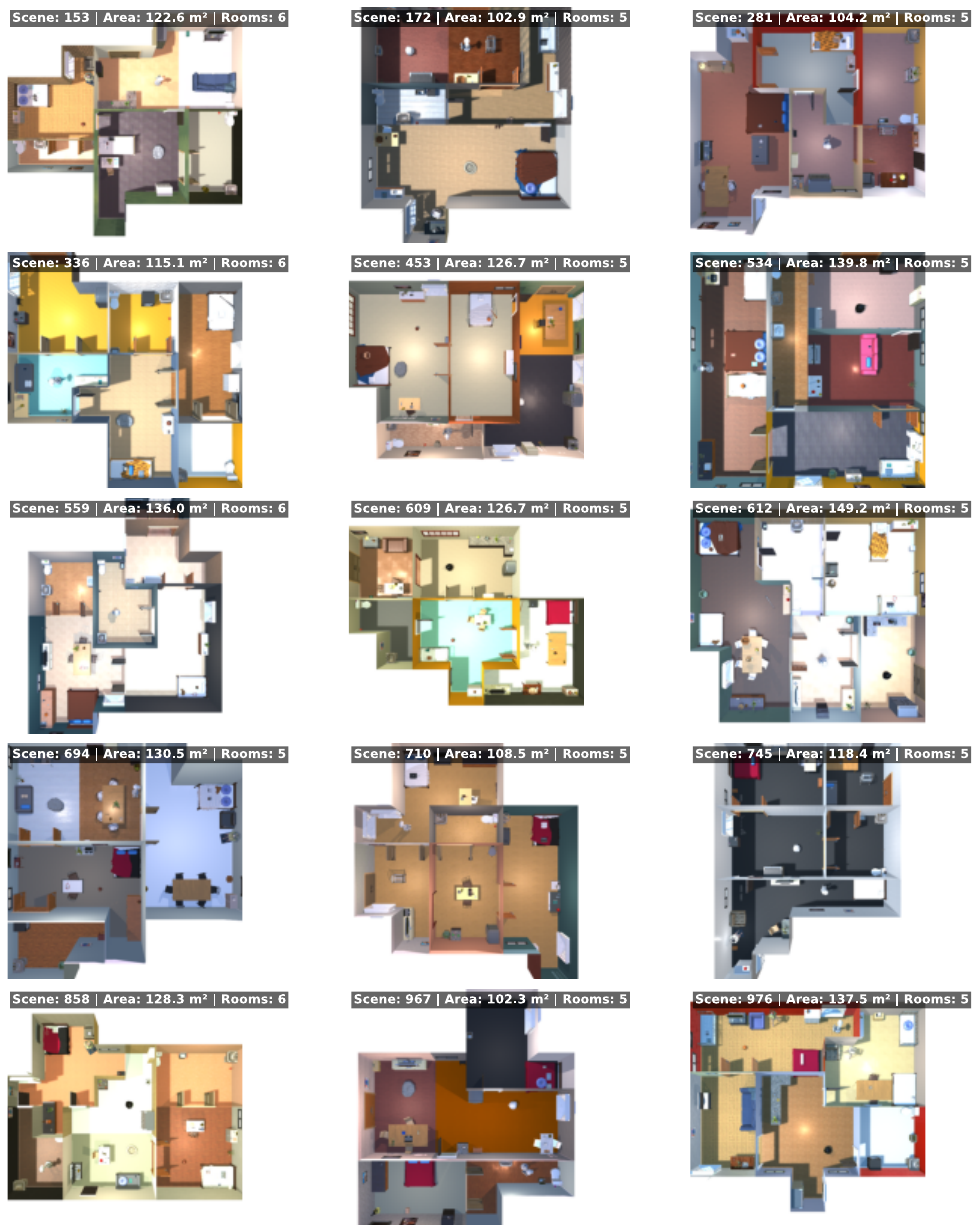}
    \caption{\textbf{ProcTHOR environments:} Top-down view of the different 15 ProcTHOR scenes used in the simulation experiments in Sec.~\ref{sec:eval_pg}. Each scene has at least 5 rooms and a navigable space between \SI{100}{\metre\squared} and \SI{150}{\metre\squared}.}
    \label{fig:procthor_environments}
\end{figure*}

\subsection{Hierarchical Scheduling Model}
\label{app:sim_schedule}
The spatial distribution of each pickupable object over time is governed by a hierarchical temporal model inspired by human activity patterns. For each pickupable, we generate a schedule that specifies which receptacle contains that object at each timestep.
This schedule uses a hierarchical structure with different temporal scales (e.g., year $\to$ month $\to$ week $\to$ day $\to$ hour) and different divisions for each hierarchical level. 
Transitions between hierarchical levels and receptacles follow a Markov process. For instance, a schedule whose hierarchical levels are day and hour, with $5$ different day-level patterns, will have a one day-level transition matrix and 5 different hour-level transition matrices. Following Kurenkov et al.~\cite{kurenkov2023modeling}, the transition matrices are defined using weights computed from the relationship between ``pickupable'' and ``receptacle'' objects defined in ProcTHOR. Duration in each receptacle is sampled from a uniform distribution, also informed by the ProcTHOR semantic prior. This hierarchical structure allows us to capture realistic patterns: objects stay in certain locations for hours or days, but may transition to different receptacles across longer timescales.

\begin{figure}[!tbp]
    \centering
    \includegraphics[width=\columnwidth]{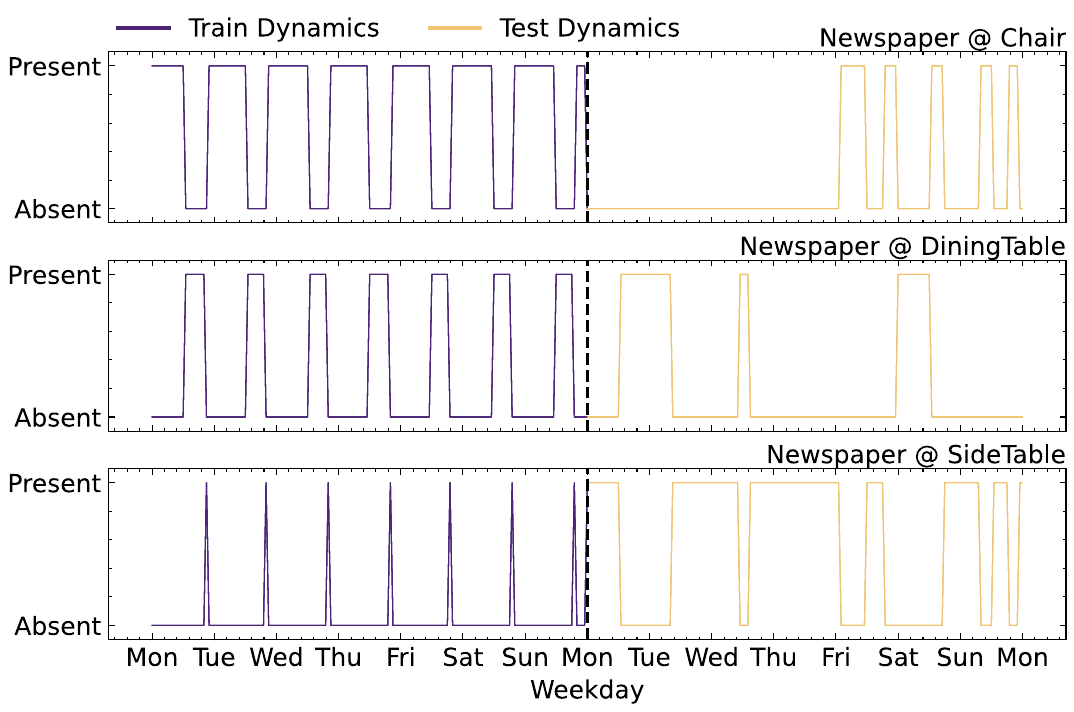}
    \caption{\textbf{Dynamic Shift in ProcTHOR Adaptation Experiments:} Illustration of the train and test time dynamics for the adaptation experiments in Sec.~\ref{sec:sim_results}. The first week displays standard training dynamics for the ``newspaper'' object across three receptacles. In the second week, these dynamics change, introducing a shift that challenges predictive methods, requiring approaches capable of real-time adaptation such as \coolPerpetua.}
    \label{fig:procthor_schedule}
\end{figure}

Figure~\ref{fig:procthor_schedule} shows a two-week generated schedule. The first week shows standard train-time dynamics, while the second week exhibits shifted test-time dynamics, as used during our adaptation experiments in Sec.~\ref{sec:sim_results}.

\subsection{Temporal Resolution}
Observations are collected at a fixed temporal $\delta t$ of 1 hour. We generate continuous trajectory data by simulating 4 weeks of time per environment, yielding a total of 672 timesteps per environment (partitioned into two weeks for training, one for validation, and one for testing). We use 15 different ProcThor scenes, which we select to have at least 5 rooms each and a navigable space between \SI{100}{\metre\squared} and \SI{150}{\metre\squared}. A top-down view of the scenes is shown in Fig.~\ref{fig:procthor_environments}.

\subsection{Relationship Between Groundtruth and Privileged Data}
\label{app:privileged} 
The groundtruth assignment indicates the simulator's ``true'' state (the receptacle that contains the object), while the privileged information filters this through the current observability constraints imposed by the camera viewpoint and rendering. Formally:

\begin{equation}
\text{privileged}_{s,r,t} = \begin{cases}
  1 & \text{if } \text{gt}_{s,r,t} = 1 \land \text{obj.~visible},\\
  0 & \text{if } \text{gt}_{s,r,t} = 0 \land \text{rec.~visible}, \\
  -1 & \text{if }   \text{rec.~hidden}, \\
  -2 & \text{if } r \text{ invalid for } s;
\end{cases}
\end{equation}

\noindent 
where \texttt{1} denotes presence, \texttt{0} absence, \texttt{-1} missing data, and \texttt{-2} that the receptacle $r$ is invalid for object $s$.

\section{Real-World Details: \coolName}
\label{app:real_details}

\begin{figure}[!tbp]
    \centering
    \includegraphics[width=\columnwidth]{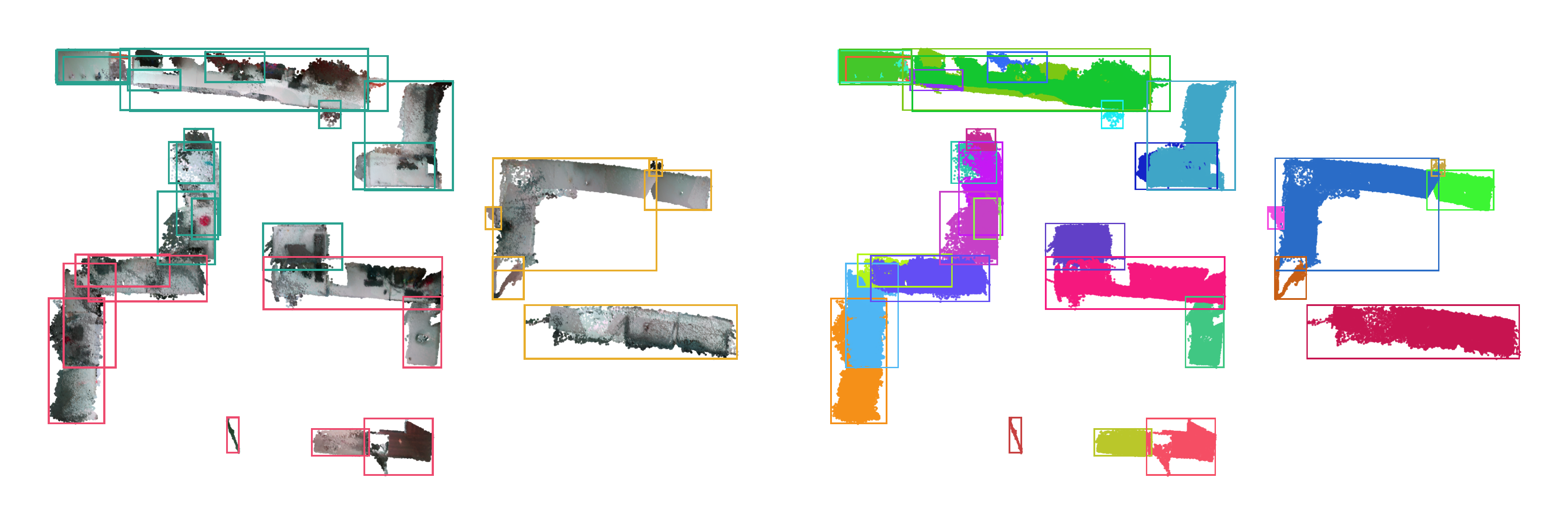}
    \caption{\textbf{Open-Vocabulary Receptacle Detection in the Real-World Experiments:} (left) colored point cloud with the bounding boxes of the detected receptacles; (right) the corresponding semantic segmentation, with all points assigned the \emph{receptacle} class. Open-vocabulary detection following DualMap~\cite{jiang2025dualmap} under-segments the receptacles in the laboratory, recovering only a subset of the valid receptacles (cf. the top panel of Fig.~\ref{fig:real_pcd}). Relying on this detector would confound prediction errors with perception errors across all methods.}
    \label{fig:under_segmentation}
\end{figure}

\subsection{Real-World Details}
\label{app:real}

\begin{figure}[!tpb]
    \centering
    \includegraphics[width=\columnwidth]{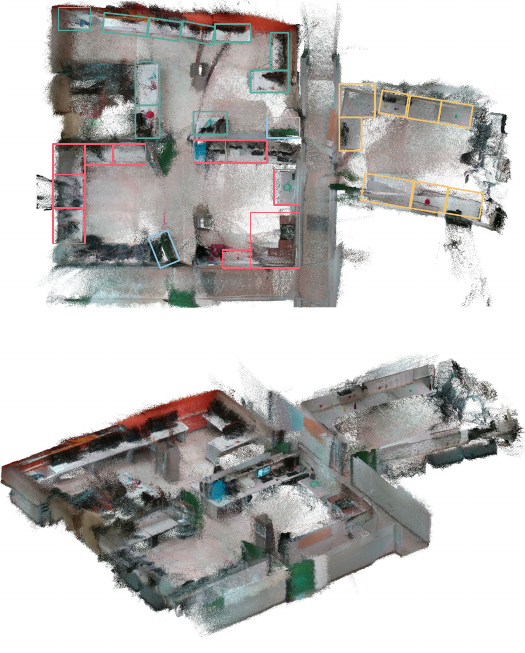}
    \caption{\textbf{Real-World Environment}: (Top) Top-down view showing receptacle and door bounding boxes. Receptacles are color-coded by room: {\color{pink}pink} (first room), {\color{teal}teal} (second room), and {\color{orange}mustard} (third room). Door bounding boxes are highlighted in {\color{light_blue}blue}. (Bottom) An isometric view of the three-room laboratory setup used for all real-world experiments.}
    \label{fig:real_pcd}
\end{figure}

For our real-world evaluation, we construct a dataset where we manually relocate objects according to a known ground-truth schedule. We deployed an Agilex Ranger Mini 2.0 robot equipped with an Intel RealSense D435 camera and teleoperated it through multiple trajectories to gather RGB-D observations. RTAB-Map~\cite{labbe2019rtab} was used as the pose estimation backend and mapping tool. To evaluate \coolName\ under realistic perception noise, we generated object-receptacle training labels for \coolPerpetua\ using a custom perception pipeline based on SAM3~\cite{carion2025sam} and the association function described in Sec.~\ref{sec:pg_method}.\looseness-1

The environment consists of a three-room laboratory connected by two doors (Fig.~\ref{fig:real_pcd}), containing a total of 28 receptacles distributed as 10, 10, and 8 across the rooms, respectively. Throughout the experiments, the two doors were periodically opened and closed to vary the environment's topological connectivity. Scans were taken bi-hourly from 9:00 AM to 5:00 PM over a span of three weeks, yielding a dataset of 84 scans. For each scan, we provide RGB-D frames and camera poses in a common origin frame, along with manually annotated 3D bounding boxes for the receptacles.

During data collection, we used the following 21 objects: \texttt{greenbowl, bell, panda, penguin, redbowl, bluecup, yellowcup, greencube, bluecube, headphones, duck, egg, carrot, hairbrush, redbottle, redmug, redplate, woodenspoon, spoon, spatula, blueball}.

\subsection{Open-Vocabulary Receptacle Detection}
\label{app:receptacle_detection}

Open-vocabulary recognition in \coolName applies to objects, which are identified without a fixed object taxonomy or manual re-annotation. Receptacles, in contrast, come from a per-scene set of named instances with 3D extents that is annotated once before deployment, and open-vocabulary object detections are matched onto this fixed set. We annotate receptacles because open-vocabulary receptacle detection under-segments our laboratory: as shown in Fig.~\ref{fig:under_segmentation}, applying the open-vocabulary pipeline of DualMap~\cite{jiang2025dualmap} recovers only a subset of the valid receptacles (cf. the top panel of Fig.~\ref{fig:real_pcd}). Since our evaluation targets the accuracy of predicted future object locations, relying on this detector would confound prediction errors with perception errors, so we use the same annotated receptacles for all methods. Receptacles are assumed to move only over long timescales, consistent with prior work~\cite{jiang2025dualmap}.

\section{Extended Navigation Results}

\begin{figure*}[!tp]
    \centering
    \includegraphics[width=\linewidth, trim=0 4.16cm 0 0cm,
    clip]{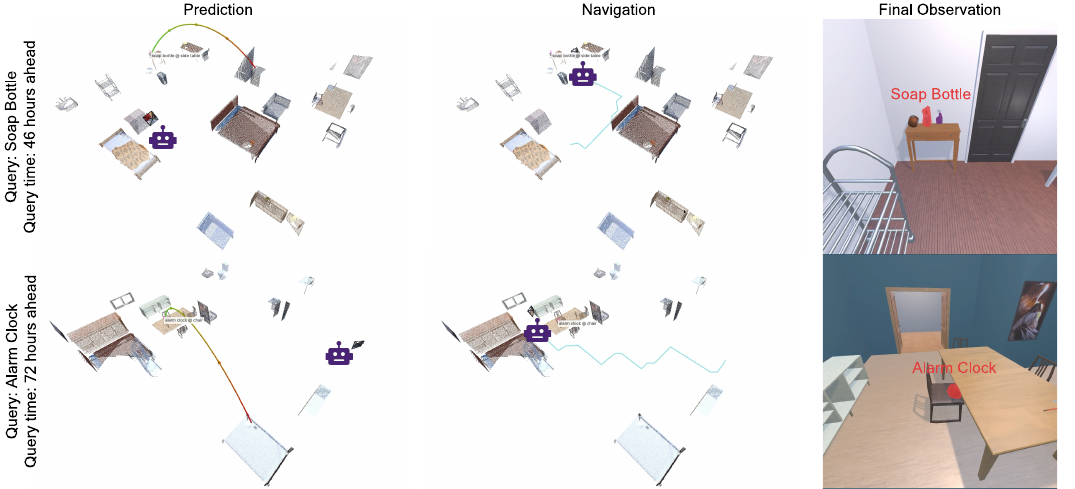}
    \caption{\textbf{Qualitative Navigation Results in ProcTHOR:} The top row shows a navigation sequence where \coolName is queried 46 hours after the last map update (top-down map shown in the first row, third column of Fig.~\ref{fig:procthor_environments}). In the first column, our method predicts that the \texttt{soap bottle} has moved from the dresser ({\color{pink}red spline tip}) to a side table ({\color{teal}green spline tip}). The second column shows the robot navigating to the predicted location, with its path marked in {\color{light_blue}blue}. The final column displays the successful detection of the soap bottle. The bottom row depicts a similar example with an \texttt{alarm clock} queried 72 hours after the last map update (top-down map shown in the third row, first column of Fig.~\ref{fig:procthor_environments}).\looseness-1}
    \label{fig:qualitative_sim}
\end{figure*}

\subsection{Simulation results}
\label{app:nav_sim_results}

\begin{table*}[htpb!]
\centering
\renewcommand{\arraystretch}{1.05}
\setlength{\tabcolsep}{3.0pt}
\caption{Adaptive Navigation Experiment: Detailed version of Fig.~\ref{fig:barplots} (left).}
\vspace{-0.5em}
\begin{tabular}{c l cc c cccc c cc}
\toprule
& & \multicolumn{2}{c}{\textbf{Static Setting} \scriptsize{(80)}}& \phantom{a}& \multicolumn{4}{c}{\textbf{Dynamic Setting} \scriptsize{(45)}}& \phantom{a}& \multicolumn{2}{c}{\textbf{Overall} \scriptsize{(125)}} \\
\cmidrule(lr){3-4} \cmidrule(lr){6-9} \cmidrule(lr){11-12}
& & \multicolumn{2}{c}{Presence \scriptsize{(80)}} && \multicolumn{2}{c}{Presence \scriptsize{(38)}} & \multicolumn{2}{c}{Absence \scriptsize{(7)}} && \multicolumn{2}{c}{Average} \\
\cmidrule(lr){3-4} \cmidrule(lr){6-7} \cmidrule(lr){8-9} \cmidrule(lr){11-12}
Perception & Method & Success (\%) $\uparrow$ & SPL (\%) $\uparrow$ && Success (\%) $\uparrow$ & SPL (\%) $\uparrow$ & \acs{AR} (\%) $\uparrow$ & Dist (m) $\downarrow$ && Success (\%) $\uparrow$ & SPL (\%) $\uparrow$ \\
\midrule
\multirow{5}{*}{Privileged}
& CG~\cite{gu2024conceptgraphs} & $52.0$\tinyvar{9.9} & $51.6$\tinyvar{9.8} && $0.0$\tinyvar{0.0} & $0.0$\tinyvar{0.0} & $66.7$\tinyvar{19.2} & \textbf{0.97}\tinyvar{0.53} & & $37.6$\tinyvar{8.1} & $36.3$\tinyvar{8.4} \\
& HOMER & $63.3$\tinyvar{2.5} & $62.7$\tinyvar{2.6} && $25.7$\tinyvar{10.3} & $24.9$\tinyvar{10.3} & $66.7$\tinyvar{33.3} & $6.0$\tinyvar{2.0} & & $53.6$\tinyvar{2.4} & $50.8$\tinyvar{2.1} \\
& FlowMaps & $61.3$\tinyvar{6.9} & $59.3$\tinyvar{6.9} && $20.3$\tinyvar{7.3} & $19.0$\tinyvar{7.0} & \textbf{88.9}\tinyvar{11.1} & $4.3$\tinyvar{2.2} & & $50.4$\tinyvar{3.2} & $46.7$\tinyvar{2.3} \\
& Ours (FreMEn~\cite{krajnik2017fremen}) & $79.3$\tinyvar{2.6} & $78.1$\tinyvar{3.0} && $53.0$\tinyvar{13.4} & $48.8$\tinyvar{11.9} & $44.4$\tinyvar{29.4} & $9.3$\tinyvar{2.7} & & $72.0$\tinyvar{3.6} & $70.2$\tinyvar{2.8} \\
& \textbf{Ours} & \textbf{89.3}\tinyvar{2.7} & \textbf{83.4}\tinyvar{1.3} && \textbf{65.4}\tinyvar{10.4} & \textbf{56.3}\tinyvar{9.9} & \textbf{88.9}\tinyvar{11.1} & $7.7$\tinyvar{2.3} & & \textbf{83.2}\tinyvar{4.3} & \textbf{75.9}\tinyvar{2.9} \\
\midrule
\multirow{5}{*}{CLIP + LLM}
& CG~\cite{gu2024conceptgraphs} & $47.4$\tinyvar{8.5} & $47.0$\tinyvar{8.4} && $0.0$\tinyvar{0.0} & $0.0$\tinyvar{0.0} & $66.7$\tinyvar{19.2} & \textbf{0.97}\tinyvar{0.53} & & $34.4$\tinyvar{7.4} & $33.1$\tinyvar{7.7} \\
& HOMER & $46.2$\tinyvar{4.8} & $45.7$\tinyvar{4.8} && $16.4$\tinyvar{4.4} & $16.4$\tinyvar{4.4} & $66.7$\tinyvar{33.3} & $7.2$\tinyvar{3.0} & & $39.2$\tinyvar{4.3} & $36.3$\tinyvar{3.0} \\
& FlowMaps & $40.5$\tinyvar{7.0} & $39.4$\tinyvar{6.9} && $9.9$\tinyvar{2.8} & $9.4$\tinyvar{2.7} & $88.9$\tinyvar{11.1} & $6.8$\tinyvar{3.7} & & $32.8$\tinyvar{5.1} & $29.3$\tinyvar{3.8} \\
& Ours (FreMEn~\cite{krajnik2017fremen}) & $57.1$\tinyvar{4.8} & $55.9$\tinyvar{4.9} && $37.2$\tinyvar{9.4} & $36.2$\tinyvar{8.7} & $44.4$\tinyvar{29.4} & $10.9$\tinyvar{3.8} & & $52.0$\tinyvar{5.7} & $50.7$\tinyvar{5.2} \\
& \textbf{Ours} & \textbf{58.5}\tinyvar{3.6} & \textbf{57.3}\tinyvar{3.8} && \textbf{38.6}\tinyvar{8.7} & \textbf{37.2}\tinyvar{8.1} & \textbf{100.0}\tinyvar{0.0} & $9.8$\tinyvar{4.3} & & \textbf{55.4}\tinyvar{5.1} & \textbf{51.6}\tinyvar{4.4} \\
\bottomrule
\end{tabular}
\begin{minipage}{0.98\linewidth}
    \footnotesize
    \vspace{0.5em}
    \textcolor{black!80}{The adaptive capabilities of \coolPerpetua enable better performance even under distributional shifts in environment dynamics. While perception errors degrade performance by introducing spurious belief updates, \coolPerpetua maintains its overall advantage due to its adaptation capabilities.\looseness-1}
\end{minipage}
\vspace{-1em}
\label{tab:adaptive_sim_full}
\end{table*}

\begin{table*}[t]
\centering
\renewcommand{\arraystretch}{1.0}
\setlength{\tabcolsep}{3.3pt}
\caption{Predictive Navigation Results}
\begin{tabular}{c l cc c ccc c cc}
\toprule
& & \multicolumn{2}{c}{\textbf{Static} - \scriptsize{Presence (157)}}& \phantom{a}& \multicolumn{2}{c}{\textbf{Dynamic} - \scriptsize{Presence (75)}} & \multicolumn{1}{c}{\textbf{Dynamic} - \scriptsize{Absence (18)}}& \phantom{a}& \multicolumn{2}{c}{\textbf{Overall} \scriptsize{(250)}} \\
\cmidrule(lr){3-4} \cmidrule(lr){6-7} \cmidrule(lr){8-8} \cmidrule(lr){10-11}
Perception & Method & Success (\%) $\uparrow$ & SPL (\%) $\uparrow$ && Success (\%) $\uparrow$ & SPL (\%) $\uparrow$ & \acs{AR} (\%) $\uparrow$ && Success (\%) $\uparrow$ & SPL (\%) $\uparrow$ \\
\midrule
\multirow{7}{*}{\rotatebox[origin=c]{90}{Privileged}}
& CG~\cite{gu2024conceptgraphs} & $51.4$\tinyvar{7.0} & $49.3$\tinyvar{6.6} && $0.0$\tinyvar{0.0} & $0.0$\tinyvar{0.0} & $46.4$\tinyvar{16.3} & & $35.2$\tinyvar{4.8} & $33.2$\tinyvar{4.9} \\
& CG + LLM & $88.1$\tinyvar{3.7} & $81.1$\tinyvar{3.5} && $68.5$\tinyvar{6.3} & $44.7$\tinyvar{6.5} & $20.2$\tinyvar{10.6} & & $76.4$\tinyvar{3.3} & $68.4$\tinyvar{3.3} \\
& CG + LLM + Hist. & $94.4$\tinyvar{2.2} & $88.5$\tinyvar{2.5} && $65.6$\tinyvar{4.8} & $41.7$\tinyvar{5.3} & $14.3$\tinyvar{9.9} & & $80.0$\tinyvar{3.0} & $73.2$\tinyvar{3.7} \\
& HOMER & $89.4$\tinyvar{4.1} & $83.9$\tinyvar{4.6} && $86.1$\tinyvar{5.5} & $69.1$\tinyvar{7.8} & $58.3$\tinyvar{17.5} & & $87.1$\tinyvar{2.9} & $80.4$\tinyvar{3.6} \\
& FlowMaps & $90.2$\tinyvar{4.0} & $85.0$\tinyvar{5.7} && $61.0$\tinyvar{5.3} & $54.1$\tinyvar{5.2} & $42.9$\tinyvar{17.4} & & $78.3$\tinyvar{3.3} & $75.9$\tinyvar{4.2} \\
& \textbf{Ours (FreMEn~\textnormal{\cite{krajnik2017fremen}})} & \textbf{97.9}\tinyvar{1.1} & \textbf{93.5}\tinyvar{1.3} & & \textbf{92.8}\tinyvar{3.1} & \textbf{81.5}\tinyvar{4.1} & \textbf{90.5}\tinyvar{6.1} & & \textbf{95.2}\tinyvar{1.4} & \textbf{88.9}\tinyvar{2.3} \\
& \textbf{Ours} & \textbf{97.9}\tinyvar{1.1} & \textbf{93.5}\tinyvar{1.3} & & \textbf{92.8}\tinyvar{3.1} & $81.1$\tinyvar{3.9} & \textbf{90.5}\tinyvar{6.1} & & \textbf{95.2}\tinyvar{1.4} & \textbf{88.9}\tinyvar{2.3} \\
\midrule
\multirow{7}{*}{\rotatebox[origin=c]{90}{CLIP + LLM}}
& CG~\cite{gu2024conceptgraphs} & $43.0$\tinyvar{6.7} & $41.0$\tinyvar{6.4} && $0.0$\tinyvar{0.0} & $0.0$\tinyvar{0.0} & $46.4$\tinyvar{16.3} & & $29.6$\tinyvar{4.5} & $27.4$\tinyvar{4.6} \\
& CG + LLM & $63.5$\tinyvar{4.9} & $56.9$\tinyvar{4.6} && $55.6$\tinyvar{6.5} & $35.2$\tinyvar{4.9} & $16.7$\tinyvar{10.9} & & $57.2$\tinyvar{4.5} & $49.7$\tinyvar{4.0} \\
& CG + LLM + Hist. & $65.3$\tinyvar{5.6} & $61.6$\tinyvar{5.7} && $51.2$\tinyvar{5.5} & $30.5$\tinyvar{4.1} & $40.5$\tinyvar{15.4} & & $58.4$\tinyvar{4.5} & $51.4$\tinyvar{4.9} \\
& HOMER & $61.9$\tinyvar{5.0} & $59.6$\tinyvar{5.3} && $71.7$\tinyvar{6.3} & $57.4$\tinyvar{6.8} & $58.3$\tinyvar{17.5} & & $65.1$\tinyvar{2.9} & $59.7$\tinyvar{3.0} \\
& FlowMaps & $62.7$\tinyvar{6.9} & $59.9$\tinyvar{7.0} && $50.0$\tinyvar{6.3} & $45.0$\tinyvar{6.6} & $42.9$\tinyvar{17.4} & & $57.5$\tinyvar{4.7} & $55.2$\tinyvar{5.4} \\
& \textbf{Ours (FreMEn~\textnormal{\cite{krajnik2017fremen}})} & \textbf{68.4}\tinyvar{4.0} & \textbf{65.2}\tinyvar{4.4} && $73.8$\tinyvar{7.6} & $65.3$\tinyvar{6.5} & \textbf{90.5}\tinyvar{6.1} & & \textbf{71.6}\tinyvar{2.6} & $65.0$\tinyvar{3.7} \\
& \textbf{Ours} & \textbf{68.4}\tinyvar{4.0} & \textbf{65.2}\tinyvar{4.4} && \textbf{76.2}\tinyvar{8.3} & \textbf{67.0}\tinyvar{7.0} & \textbf{90.5}\tinyvar{6.1} & & \textbf{71.6}\tinyvar{2.9} & \textbf{65.1}\tinyvar{3.1} \\
\bottomrule
\end{tabular}
\label{tab:predictive_sim}
\begin{minipage}{0.98\textwidth}
    \footnotesize
    \vspace{0.5em}
    \textcolor{black!80}{\coolName with \coolPerpetua or FreMEn estimators outperforms baselines under both privileged and noisy perception, including HOMER and FlowMaps baselines.}
\end{minipage}
\vspace{-1.2em}
\end{table*}

Table~\ref{tab:adaptive_sim_full} presents the comprehensive adaptive navigation results, expanding on the results provided in Sec.~\ref{sec:sim_results}. Here, we ablate \coolName by comparing the fully-predictive FreMEn estimator against our adaptive \coolPerpetua method. Under privileged perception, \coolName with \coolPerpetua achieves superior navigation performance, effectively adapting to the shift in test-time dynamics. Even when perception errors degrade performance, particularly in the dynamic setting, \coolPerpetua maintains its advantage over the FreMEn variant. As demonstrated by the privileged results, we expect that as perception pipelines improve, the performance of \coolName\ will scale accordingly.

Additionally, Fig.~\ref{fig:qualitative_sim} presents qualitative results for the ProcTHOR navigation experiments from Sec.~\ref{sec:sim_results}. This figure shows how \coolName accurately predicts the location of target objects at different query times. This capability allows the agent to exploit predictive information to successfully navigate to its target, even when the underlying scene graph has not been updated for several hours.

In Table~\ref{tab:predictive_sim} we provide the same results shown in Fig.~\ref{fig:main_batplot} in table form.

\begin{figure}[!tbp]
    \centering
    \includegraphics[width=\columnwidth]{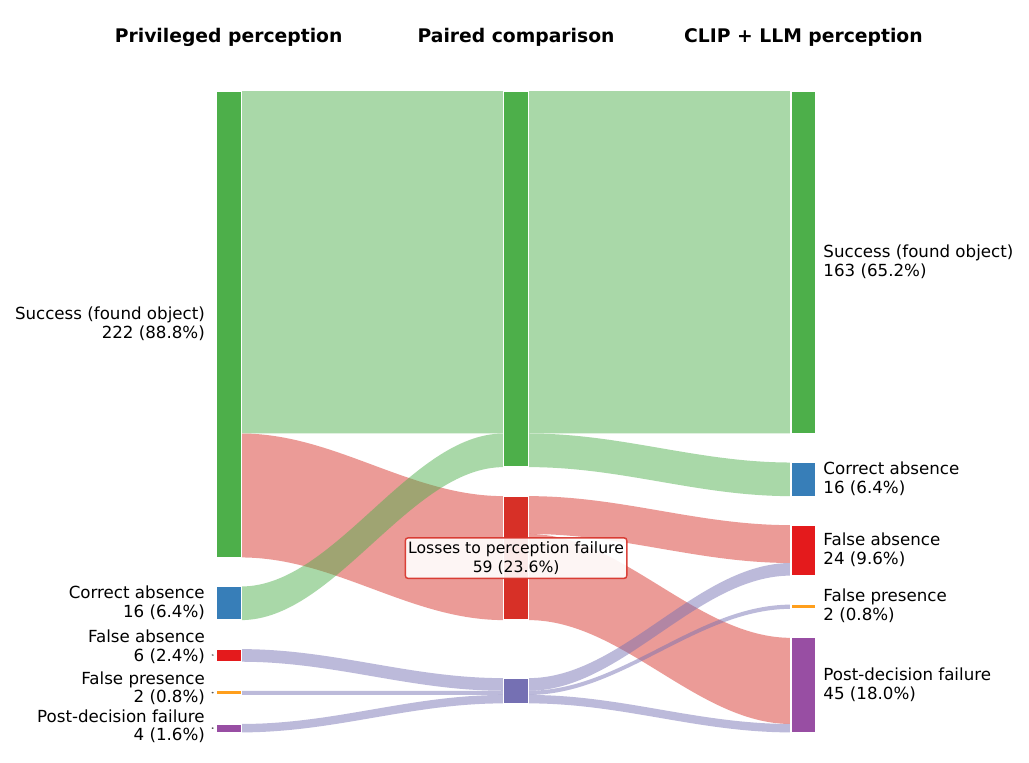}
    \caption{\textbf{Failure Taxonomy:} paired Sankey diagram over the 250 predictive navigation queries in simulation (10 ProcTHOR layouts $\times$ 25 queries). Each flow traces one query from its outcome under privileged perception (left), through a paired comparison (middle), to its outcome under CLIP + LLM perception (right). The net gap induced by perception is 23.6\%.}
    \label{fig:failure_sankey}
\end{figure}

To categorize failures, Fig.~\ref{fig:failure_sankey} traces each of the 250 predictive navigation queries (10 ProcTHOR layouts $\times$ 25 queries) from its outcome under privileged perception to its outcome under CLIP + LLM perception for \coolName. Moving from privileged to noisy perception, the share of queries where the object is found drops from 88.8\% to 65.2\%, a net gap of 23.6 percentage points, while correct absences remain at 6.4\%. The losses appear mainly as false absences (2.4\% to 9.6\%) and post-decision failures (1.6\% to 18.0\%), while false presences are unchanged (0.8\%). We interpret this as evidence that the third-party perception system upstream of our method is a major bottleneck: as robotic perception improves, the performance of \coolName will directly improve. 

\subsection{Real-World results}
\label{app:nav_real_results}

\begin{table}[t]
\centering
\small
\renewcommand{\arraystretch}{1.0}
\setlength{\tabcolsep}{2.5pt}
\caption{\textbf{Real-World Navigation:} \coolName enables robust dynamic navigation amidst perception errors, even when trained on noisy real-world data.}
\resizebox{\columnwidth}{!}{%
\begin{tabular}{l c c c c}
\toprule
& \textbf{Static} \scriptsize{(40)} & \textbf{Dynamic} \scriptsize{(40)} & \textbf{Absent} \scriptsize{(20)} & \textbf{Overall} \scriptsize{(100)} \\
\cmidrule(lr){2-5}
Estimator & SR (\%) $\uparrow$  & SR (\%) $\uparrow$  & \acs{AR} (\%) $\uparrow$ & SR (\%) $\uparrow$ \\
\midrule
Ours (noisy oracle) & $85.0$ & $77.5$ & $100.0$ & $85.0$ \\
\midrule
CG \cite{gu2024conceptgraphs} & $\mathbf{77.5}$ & $0.0$ & $50.0$ & $41.0$\\
DualMap \cite{jiang2025dualmap} & $62.5$ & $35.0$ & $35.0$ & $46.0$\\
HOMER & $45.0$ & $60.0$ & $65.0$ & $55.0$\\
Ours & $70.0$ & $\mathbf{65.0}$ & $\mathbf{75.0}$ & $\mathbf{69.0}$\\
\bottomrule
\end{tabular}%
}
\label{tab:adaptive_real_world}
\vspace{-1em}
\end{table}

\begin{figure}[!tbp]
    \centering
    \includegraphics[width=\columnwidth]{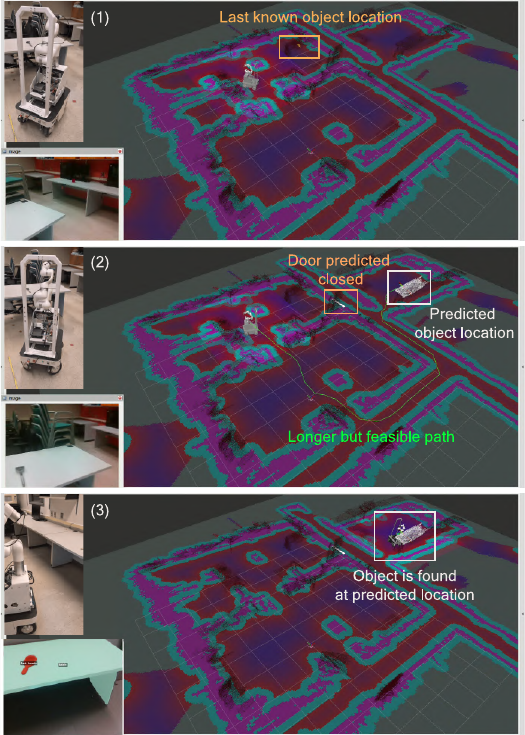}
    \caption{\textbf{Qualitative Navigation Result in Real-World - Topological Graph Change:} Extended version of Fig.~\ref{fig:doors}. The open-vocabulary query is ``\texttt{Is there something useful to comb my hair?}'', with query time 146 hours after the last map update. (1) Shows the last known position of the relevant object identified by \coolName (a hair brush). (2) Our method predicts that the hair brush is in another room and detects that the door along the shortest path is closed; this updates the cost map, prompting the planner to generate a longer but feasible path. (3) The agent successfully locates the target object.
    \looseness-1}
    \label{fig:qualitative_doors}
\end{figure}

\begin{figure}[!tbp]
    \centering
    \includegraphics[width=\columnwidth]{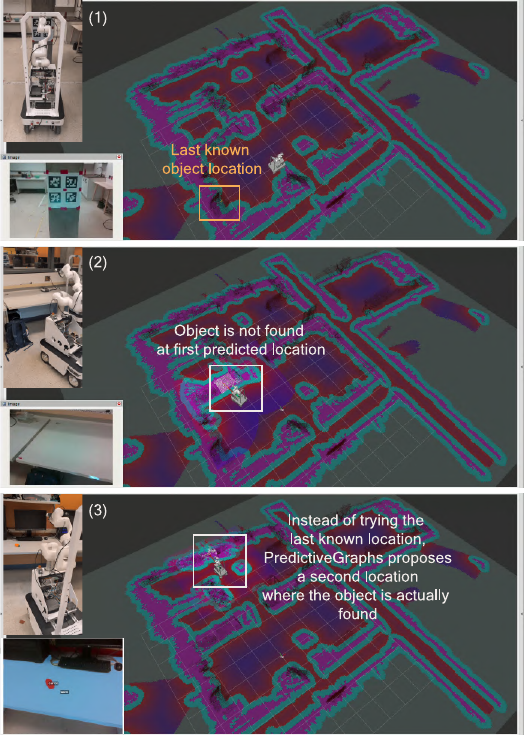}
    \caption{\textbf{Qualitative Navigation Result in Real-World - Replanning:} Open-vocabulary query ``\texttt{Could you get me a healthy vegetable?}'' with query time 29 hours after the last map update. (1) \coolName identifies the target object as a carrot. (2) \coolName proposes an initial candidate location; however, upon navigating to it, the object is not found. (3) Instead of reverting to the object's last known location, \coolName predicts a new location where the carrot is successfully found, highlighting the robustness of our tempo-spatio-semantic scene graph in dynamic scenes.}
    \label{fig:qualitative_replanning}
    \vspace{-1em}
\end{figure}

In this section, we provide additional qualitative results to supplement the real-world navigation experiments discussed in Sec.~\ref{sec:real_world_results}. Figure~\ref{fig:qualitative_doors} presents an extended version of Fig.~\ref{fig:doors}, demonstrating the Agilex Ranger Mini 2.0 successfully navigating to its target, a hair brush. By anticipating that the shortest path is blocked, our method proactively plans a longer, yet feasible, alternative route to circumvent the obstacle. 

Similarly, Fig.~\ref{fig:qualitative_replanning} illustrates a search for a ``healthy vegetable'' where the initial predicted location is empty. Instead of relying on the outdated map, \coolName correctly predicts a new location for the object, allowing the agent to recover and successfully complete the task. 

These experiments are also shown in the first section of the video available on the project webpage\footref{fn:supp}.

In Table~\ref{tab:adaptive_real_world} we provide the same results shown in Fig.~\ref{fig:barplots} in table form.

\end{document}